\documentclass[10pt,twocolumn,letterpaper]{article}

\usepackage[pagenumbers]{cvpr} 

\usepackage{amsmath,amsfonts,bm}

\def\eqref#1{equation~\ref{#1}}

\def\1{\bm{1}}

\DeclareMathAlphabet{\mathsfit}{\encodingdefault}{\sfdefault}{m}{sl}
\SetMathAlphabet{\mathsfit}{bold}{\encodingdefault}{\sfdefault}{bx}{n}

\usepackage{url}
\usepackage{graphicx}
\usepackage{amsmath}
\usepackage{amssymb}
\usepackage{booktabs}
\usepackage{listings}
\usepackage{float}
\usepackage{listings}
\usepackage{tcolorbox}
\usepackage{adjustbox}

\definecolor{myblue}{rgb}{0.07, 0.39, 0.49}
\definecolor{mygreen}{rgb}{0.12, 0.67, 0.23}
\definecolor{tan}{rgb}{0.93, 0.93, 0.88}

\usepackage{wrapfig}

\usepackage[pagebackref,breaklinks,colorlinks]{hyperref}

\usepackage[capitalize]{cleveref}
\crefname{section}{Sec.}{Secs.}
\Crefname{section}{Section}{Sections}
\Crefname{table}{Table}{Tables}
\crefname{table}{Tab.}{Tabs.}

\def\cvprPaperID{14714} 
\def\confName{CVPR}
\def\confYear{2025}

\begin{document}

\title{Harnessing Frozen Unimodal Encoders for Flexible Multimodal Alignment}

\author{Mayug Maniparambil\thanks{joint first authors} \hspace{0.5em}\thanks{ML Labs, Dublin City University} \and Raiymbek Akshulakov\footnotemark[1]\hspace{0.5em}\thanks{University of California Berkeley} \and Yasser Abdelaziz Dahou Djilali\thanks{Technological Innovation Institute}\hspace{0.5em}\footnotemark[2] \and Sanath Narayan\footnotemark[4] \and Ankit Singh\footnotemark[4] \and Noel E. O'Connor\footnotemark[2]}
\maketitle

\begin{abstract}

   Recent contrastive multimodal vision-language models like CLIP have demonstrated robust open-world semantic understanding, becoming the standard image backbones for vision-language applications. However, recent findings suggest high semantic similarity between well-trained unimodal encoders, which raises a key question: Is there a plausible way to connect unimodal backbones for vision-language tasks? To this end, we propose a novel framework that aligns vision and language using frozen unimodal encoders. It involves selecting semantically similar encoders in the latent space, curating a concept-rich dataset of image-caption pairs, and training simple MLP projectors. We evaluated our approach on 12 zero-shot classification datasets and 2 image-text retrieval datasets. Our best model, utilizing DINOv2 and All-Roberta-Large text encoder, achieves 76\(\%\) accuracy on ImageNet with a 20-fold reduction in data and 65-fold reduction in compute requirements compared multi-modal alignment where models are trained from scratch. The proposed framework enhances the accessibility of multimodal model development while enabling flexible adaptation across diverse scenarios. Code and curated datasets are available at \texttt{github.com/mayug/freeze-align}.
\end{abstract}

\section{Introduction}

Contrastive multimodal vision-language models have recently demonstrated impressive zero-shot capabilities \cite{radford2021learning, jia2021scaling, zhai2023sigmoid}. These advancements facilitate the use of language as an API for vision tasks, treating captions as adaptive classes to support a wide range of applications. However, current models face significant challenges: the typical objective function, InfoNCE, is designed to maximize mutual information between the global summary vector of an image and its text representation. This global approach, which relies on pooling functions within the CLIP vision encoder, struggles to deliver the pixel-level granularity required for tasks like segmentation \cite{bicaimproving}. In contrast, recent advances in uni-modal vision encoders, such as the DINOv2  \cite{oquab2023dinov2}, have demonstrated strong performance in both global and local vision tasks.  The CLIP text encoder is limited by its English-only tokenizer and a fixed token length of 77, restricting it's long-context and multilingual retrieval capabilities.
Meanwhile unimodal language encoders \cite{reimers2024sentence-transformers}, excel in multilingual, and long-context abilities, as evidenced by improved performance on MTEB benchmarks \cite{muennighoff2022mteb}. Despite these advances in unimodal models, the current strategy for aligning vision and language models usually involves full retraining of vision and language encoders, which is both computationally expensive and inflexible.

This paper proposes a framework for vision-language alignment that efficiently leverages advanced uni-modal vision and language encoders, creating adaptable multimodal models by training only projectors between their frozen embedding spaces. Current efforts to create more efficient CLIP models often compromise on either performance or still require significant resources. For example, LiT \cite{zhai2022lit} achieves comparable results to CLIP but relies on massive compute resources, while smaller-scale models like LiLT \cite{khancontrastive} may lack sufficient concepts in their training datasets, limiting their zero-shot domain transfer accuracy.

To address these challenges, our approach builds on recent findings suggesting semantic similarities between well-trained unimodal vision and language embedding spaces \cite{Maniparambil_2024_CVPR, huh2024platonic}. We hypothesize that these similarities enable effective alignment through simple projection transformations, and verify through a toy example in Section \ref{sec:cka_toy_example} and extensive ablation studies in Section \ref{sec:cka_ablation}. Inspired by this, our framework includes three key steps: \textit{identifying semantically similar vision-language encoder pairs, curating concept-dense datasets, and training lightweight projectors for efficient alignment}.


This approach has three practical benefits compared to CLIP-like training:

\textbf{Strong Unimodal Features lead to Strong Multi-Modal Models} Features from uni-modal vision and text encoders are more general than multi-modal trained encoders. For example, it's been shown that vision-only trained encoders perform better on vision-centric tasks when compared to multi-modal vision encoders like CLIP-vision \cite{tong2024cambrian}. Hence by keeping these uni-modal encoders frozen and training only projectors for alignment, we aim to keep these strong uni-modal features intact, resulting in better multi-modal representations(See Sec. \ref{sec:localization}).
\textbf{Flexible adaptation to diverse scenarios:} By utilizing the frozen unimodal encoders ability to handle a specific type of data we can efficiently train multimodal models that also can handle this specialized data without the need to retrain the whole network from scratch. For example, multilingual or long context vision-language models can be achieved by aligning DINOv2 with a multilingual (Section \ref{subsec:multiling}) or long-context language text encoder(Section \ref{subsec:dci}).
\textbf{Accessible development and Model Reuse:} Relying on already established encoders, projection heads with a dense dataset require significantly less computational resources compared to full model training. In purely practical sense, this approach not only decreases the environmental impact of developing multimodal models but also makes their creation more accessible to the broader research community  (Section \ref{subsec:compute}).

Finally, we evaluate our approach on zero-shot transfer to 12 different classification datasets and 2 image-text retrieval datasets. Our best projector between unimodal models, utilizing DINOv2 and All-Roberta-Large-v1, achieves 76\% accuracy on ImageNet, surpassing CLIP's performance while using approximately 20 times less data and 65 times less compute for alignment. We also demonstrate our framework's versatility across tasks like zero-shot domain transfer, multilingual classification, zero-shot semantic segmentation, and image-paragraph retrieval.

Our main contributions lie not in a specific model, but in demonstrating a new framework for vision-language alignment. In summary, we demonstrate that CLIP-like performance can be achieved by training only projection layers, using a curated, concept-rich dataset to enable efficient projector training with significantly less data and compute.

\begin{figure}[htbp]
\centering
  \includegraphics[width=\linewidth]{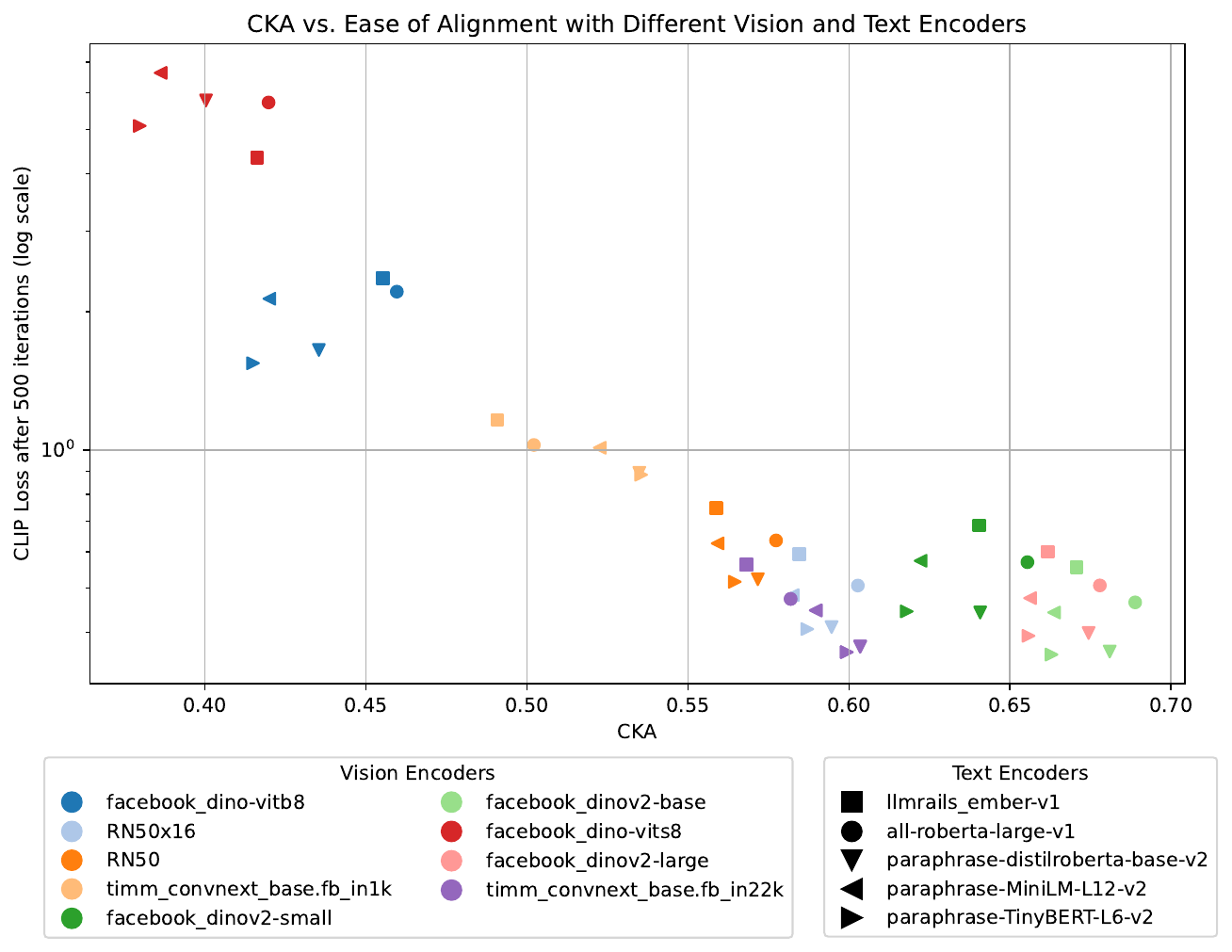}
  \vspace{-10pt}
  \setlength{\belowcaptionskip}{-5pt}
  \caption{\textbf{CLIP Loss minima vs CKA for different encoder pairs on a toy image, caption pair dataset}. We plot the CLIP loss after 500 iterations vs CKA for different image, text encoders and find that a negative correlation exists between CKA and ease of alignment.}
  \label{fig:cka_vs_clip_toy_encoder}
\setlength{\belowcaptionskip}{-10pt}

\end{figure}

\section{Related Works}

\textbf{Multimodal Pretraining:} The CLIP models from OpenAI \cite{radford2021learning} and ALIGN \cite{jia2021scaling} pioneered using web-scale image-caption data to align image and text modalities via an InfoNCE \cite{oord2018representation} loss, optimizing mutual information between embeddings. LAION \cite{schuhmann2021laion, schuhmann2022laion} replicated this approach in the open domain, open-sourcing pre-training datasets. While these models excel in zero-shot tasks, they demand substantial computational resources, around 20k GPU hours. 
Taking advantage of the recent improvements in the representation quality of unimodal encoders such as DINOv2 \cite{oquab2023dinov2} (vision) and Sentence Transformer \cite{sentencetransformer} (language) models, \cite{zhai2022lit} reduce the training cost by locking the image encoder and training only the text encoder to achieve competitive performance. Similarly, \cite{khancontrastive} further aligned frozen uni-modal encoders using projection layers, BitFit \cite{zaken2021bitfit}, and trainable adapters, but their approach is sub-optimal compared to CLIP, likely due to smaller datasets used and random encoder pair selection. In contrast, we strive to identify the best encoder pairs for alignment first and then scale up projector-only training to improve the multimodal alignment.

\textbf{Representational Similarity:} Recent studies show that the semantic similarity between vision and language model embeddings is high for several model pairs. \cite{Maniparambil_2024_CVPR} reports that this similarity, measured by Centered Kernel Alignment \cite{kornblith2019similarity}, increases with more training data for vision models. Similarly, \cite{huh2024platonic} finds that better-performing language models have higher semantic similarity to the DINOv2 \cite{oquab2023dinov2} vision model. These similarities have been leveraged for 0-shot and multi-lingual retrieval tasks using strong uni-modal encoders without additional training \cite{Maniparambil_2024_CVPR, moschella2022relative}, though scalability is an issue. Additionally, \cite{merullo2022linearly} demonstrates that a simple linear mapping allows a frozen language model to interpret visual input, provided the visual encoder aligns with language concepts (e.g., CLIP). These findings suggest that a simple projection transformation separates the embedding spaces of well-trained vision and language models, motivating our work on developing CLIP models using projection layers between semantically similar encoder pairs. 

\textbf{Automatic Data Curation:} Our dataset curation pipeline draws on various approaches in Vision-Language dataset construction \cite{radford2021learning, gadre2024datacomp, xu2024demystifying}. \cite{radford2021learning} used image metadata to gather high-quality image-caption pairs, while \cite{schuhmann2021laion} replicated the CLIP dataset by filtering with pretrained vision encoders. Recent methods like \cite{gadre2024datacomp} employ CLIP-based filtering and ad hoc filtering techniques, and \cite{xu2024demystifying} mimics CLIP’s data collection via metadata retrieval. Similarly, \cite{oquab2023dinov2} uses a pretrained vision encoder to curate web images most similar to images in curated datasets. Our approach is similar, constructing concept image prototypes from few-shot labeled examples and retrieving relevant web images from the LAION-400M pool using CLIP caption embeddings, avoiding the computational cost of generating vision embeddings for the entire dataset.

\section{CKA vs Ease of Alignment}
\label{sec:cka_vs_alignment}
Previous studies \cite{huh2024platonic, Maniparambil_2024_CVPR} have shown that well-trained vision and language encoders exhibit high semantic similarity using metrics like Centered Kernel Alignment. Specifically, a layerwise analysis in \cite{Maniparambil_2024_CVPR} reveals that most of this similarity is concentrated in the final projection layer. Furthermore, model stitching methods \cite{lenc2015understanding, bansal2021revisiting, merullo2022linearly} demonstrate that different network regions can be stitched together using linear layers suggesting that deep representations that contain high-level semantics can be connected by simple transformations. Inspired by this, we investigate whether semantically similar embedding spaces can be aligned through a simple projection transformation, using a toy example to validate the underlying concept. 

\subsection{CKA Preliminary}
\label{sec:cka_prelim}
\textbf{Centered Kernel Alignment (CKA)} has shown its relevance
in understanding and comparing the information encoded
by different layers of a neural network. CKA can be defined as follows:
Given two sets of vectors \( X \) and \( Y \), CKA measures the similarity of these vectors in their respective high-dimensional feature spaces. The kernel matrices \( K \) and \( L \) are derived from the data sets \( X \) and \( Y \), respectively, and represent the inner products between the vectors in these spaces. The entries of \( K \) and \( L \) are:
\[
K_{ij} = k(\mathbf{x}_i, \mathbf{x}_j), \quad L_{ij} = l(\mathbf{y}_i, \mathbf{y}_j)
\]
where \( k \) and \( l \) are kernel functions applied to the vectors \(\mathbf{x}_i, \mathbf{x}_j \in X\) and \(\mathbf{y}_i, \mathbf{y}_j \in Y\), respectively. Common choices for these kernel functions include linear kernels, where \( k(\mathbf{x}_i, \mathbf{x}_j) = \mathbf{x}_i^\top \mathbf{x}_j \), or Gaussian kernels, where \( k(\mathbf{x}_i, \mathbf{x}_j) = \exp(-\gamma \|\mathbf{x}_i - \mathbf{x}_j\|^2) \) for some \(\gamma > 0\).

The CKA coefficient, \( \text{CKA}(K, L) \), is defined as:
\[
\text{CKA}(K, L) = \frac{\text{HSIC}(K, L)}{\sqrt{\text{HSIC}(K, K) \cdot \text{HSIC}(L, L)}}
\]
where \( \text{HSIC} \) stands for Hilbert-Schmidt Independence Criterion \cite{gretton2005measuring, ma2020hsic}, which measures the dependence between the sets of vectors. This measure is invariant to orthogonal transformations and isotropic scaling of the data, making it robust for comparing different models.

\subsection{CKA and Ease of Alignment Toy Example}
\label{sec:cka_toy_example}

We define the \textit{Ease of Alignment} as the minimum training loss achieved after convergence, reflecting the efficiency of aligning encoder outputs. We explore how Centered Kernel Alignment (CKA) correlates with the minimum CLIP loss when transforming one vector set to match another using a Linear layer. Given the lack of a closed-form solution for CLIP loss, we employ Stochastic Gradient Descent (SGD) for $500$ iterations per instance, recording the final loss as the minimum. We fixed the temperature at $0.07$ and the learning rate at $0.01$, selecting $500$ iterations as the loss plateaued beyond this point.

In this experiment, we examine if there is an inverse relationship between the minima of CLIP Loss and CKA for embeddings derived from real data using different language and vision encoders. We sample $5000$ image-caption pairs from the COCO validation set and process them through five different sentence encoders and nine vision encoders, generating $45$ unique sets of embeddings (A and B). We calculate CKA and record the CLIP Loss after $500$ iterations for each set, plotting these values in Figure \ref{fig:cka_vs_clip_toy_encoder} with CKA on the x-axis and minima of CLIP loss on the y-axis on a log scale. The results confirm a strong inverse relationship between CKA and the minima of CLIP loss, suggesting that high CKA scores indicate similar structural similarities in encoders, which facilitate their alignment through simple projection methods. Further details on toy examples and visualization of similarity structures can be found in Sections \ref{sec:toy_example_synthetic} and \ref{sec:graph_structures}.

\section{Framework}
Our framework consists of three main components: (1) Encoder Pair Selection, (2) Dataset Curation, and (3) Lightweight Projector Training.

\subsection{Encoder Pair Selection}
 Inspired by Section \ref{sec:cka_vs_alignment} we use CKA for selecting the most semantically similar encoder pairs for multimodal alignment. We opted for a linear kernel in the CKA computation after observing that the trends in results were largely consistent between linear and RBF kernels, while the linear kernel offers superior computational efficiency. We measure the CKA between encoder spaces by constructing sets of vision embeddings and text embeddings on the COCO validation set of 5000 image, caption pairs. The COCO validation set is chosen as the reference set for its high semantic alignment between the image content and the caption description. We ablate the use of CKA for encoder pair selection in \ref{sec:cka_ablation} and find a positive correlation between CKA and transfer performance to downstream datasets.

\subsection{Dataset Curation}
\begin{figure}
    \centering
    \vspace{-10pt}
    \includegraphics[width=\linewidth, trim=0 20pt 0 0, clip]{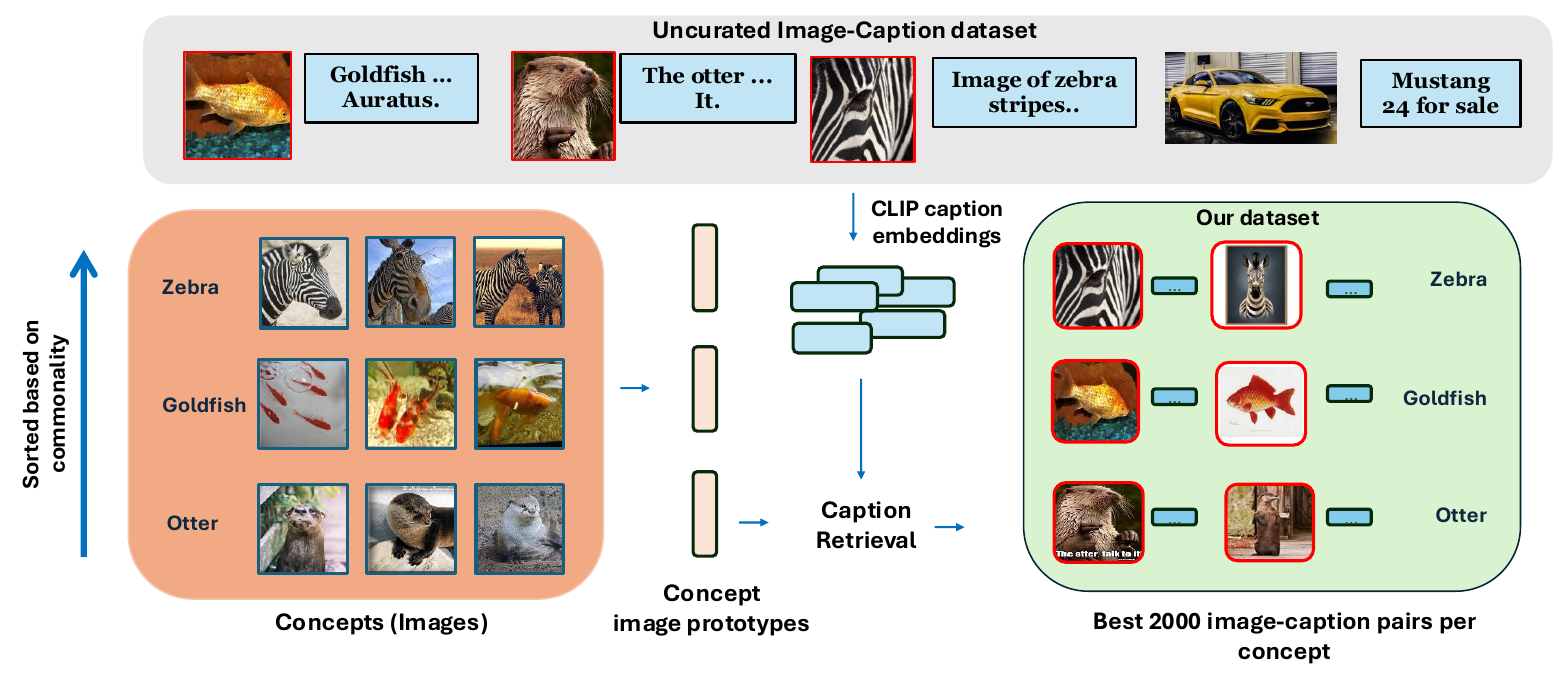}
    \setlength{\belowcaptionskip}{-15pt}
    \vspace{-10pt}
    \caption{\textbf{Overview of our concept-balanced dataset curation process.} Images for each concept are acquired from curated datasets and mapped to CLIP embeddings and averaged to construct Image Prototypes for each concept. Captions of the uncurated dataset are mapped to CLIP's joint embedding space and 2000 samples are picked per concept on the basis of the closest caption embeddings to each concept image prototype.}
    \label{fig:collection}
\end{figure}
By only training the projection layers (11M parameters) to align embedding spaces, our approach requires significantly less data compared to training a CLIP model from scratch. However, to ensure high-quality alignment and effective transfer to diverse downstream tasks, it is essential to use a small but well-curated dataset that has the following features. 1. high concept coverage which aids in covering all regions of the uni-modal embedding spaces 2.  high semantic alignment between image-caption pairs which aids in learning an effective mapping between vision and the embedding spaces. With these requirements in mind, our dataset curation process is structured into two key steps:

\textbf{Concept Coverage Collection}: 
To ensure high concept coverage, we collect $\sim$ 3000 unique concepts from class names of ImageNet, and several other curated datasets (see \ref{subsubsec:concept_datasets}).  Concept image prototypes are then constructed by averaging few-shot image embeddings for each concept using CLIP VIT-Large's vision encoder. To create a class-balanced dataset, we first collect image-caption pairs from LAION400M, a large, uncurated source dataset. We then embed all captions using CLIP ViT-Large’s text encoder and compute the caption-image prototype similarity for each concept. To ensure diversity, we retrieve 2,000 samples per concept, starting with the less common concepts. As a proxy to establish the commonality of a concept in the pool, we use the average cosine similarity of the top 25,000 captions closest to each concept prototype. This process results in LAION-CLASS-Collected, a high-quality dataset of 6M samples with broad concept coverage. The detailed algorithm is illustrated in Fig \ref{fig:collection}. \ref{sec:curation_impl} details the implementation and compute requirements for our collection process. 

Our primary goal is to compile a concept-rich dataset that enables quick learning and validates the efficacy of projectors for modality alignment, rather than developing a specific curation method. This paper demonstrates the potential of such multimodal models, emphasizing their practicality and efficiency when supported by a dataset with ample concept coverage and robust semantic alignment. The development of an exhaustive dataset that spans all domains of unimodal spaces, ensuring optimal semantic alignment between images and captions, is reserved for future work.


\textbf{Retrieval Datasets}: The LAION-CLASS-Collected dataset offers high concept diversity, but LAION itself is uncurated, with many captions poorly aligned with their images \cite{fan2024improving, nguyen2024improving, chen2023sharegpt4v}. While concept coverage is crucial for a dense coverage of the unimodal embedding space, image quality, text diversity, and image-caption alignment are key for effective zero-shot image-text retrieval. In contrast, datasets like CC3M \cite{sharma2018conceptual}, CC12M \cite{changpinyo2021conceptual}, and SBU \cite{ordonez2011im2text} feature higher-quality images and better image-caption alignment than LAION. By combining these, we create a 20M MIX-CLASS-Collected dataset that balances concept coverage with image-text similarity, resulting in both dense coverage of the uni-modal embedding spaces as well as high semantic alignment between cross-modal embeddings. We examine the impact of each data source on task performances in Sec \ref{sec:dataset_ablation}.

\subsection{Projector Architecture}
\label{sec:projector_arch}

\begin{figure}
    \centering
    \vspace{-10pt}
    \includegraphics[width=\linewidth]{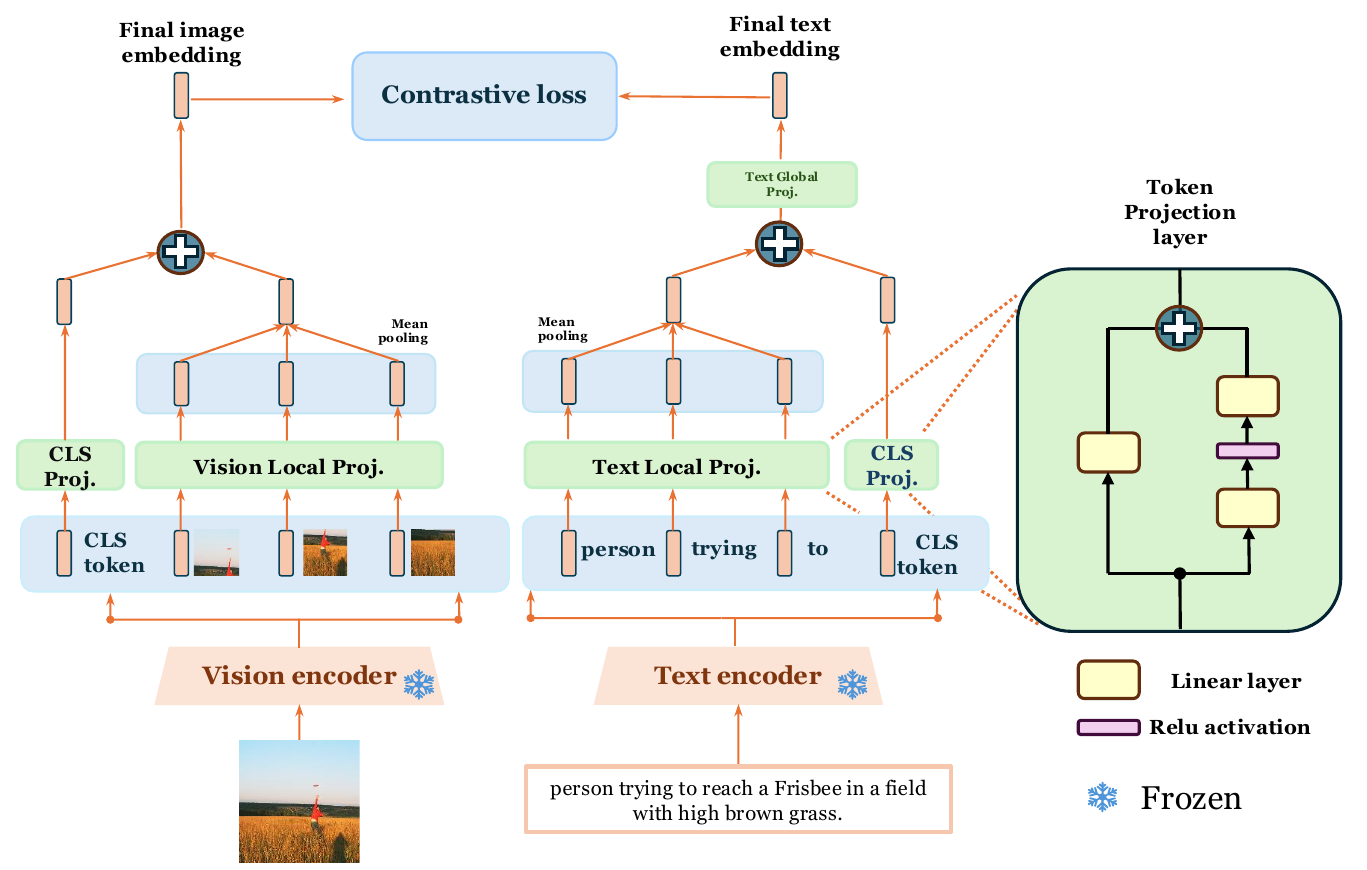}
    \caption{\textbf{Lightweight Projector Architecture.} We train only Projection Layers to align modalities. Separate projectors are applied on both the local tokens and the CLS token for each encoder and then combined in a residual manner.}
    \label{fig:proj_arc}
    \vspace{-15pt}
\end{figure}

We train lightweight projectors using contrastive loss between adapted image and text embeddings while keeping the unimodal encoders frozen. Figure \ref{fig:proj_arc} shows our projector architecture/configuration. We use a lightweight Token Projector \cite{Mukhoti_2023_CVPR} with linear and non-linear branches in a residual configuration for both local tokens and the CLS token of each encoder. The projector’s weights are shared for local tokens and separate for the CLS token to enable adaptation of both spatial and global information of the vision encoder  while limiting the parameter count. Adapted local tokens are averaged and added to the adapted CLS token to form a global embedding, capturing both global and local encoder information. For text encoders, Token Projectors are applied to the tokens, followed by a 2-layer MLP as a global Text Projector, as the text embeddings need further adaptation to become more aligned with the vision embeddings. All projector choices are thoroughly ablated in Section \ref{sec:projector_ablation}. Training information and hyperparameters are detailed in\ref{sec:projector_training}.

\begin{figure*}[!t]
    \centering
    \begin{minipage}[l]{0.6\linewidth}
        \centering
        \includegraphics[width=\linewidth]{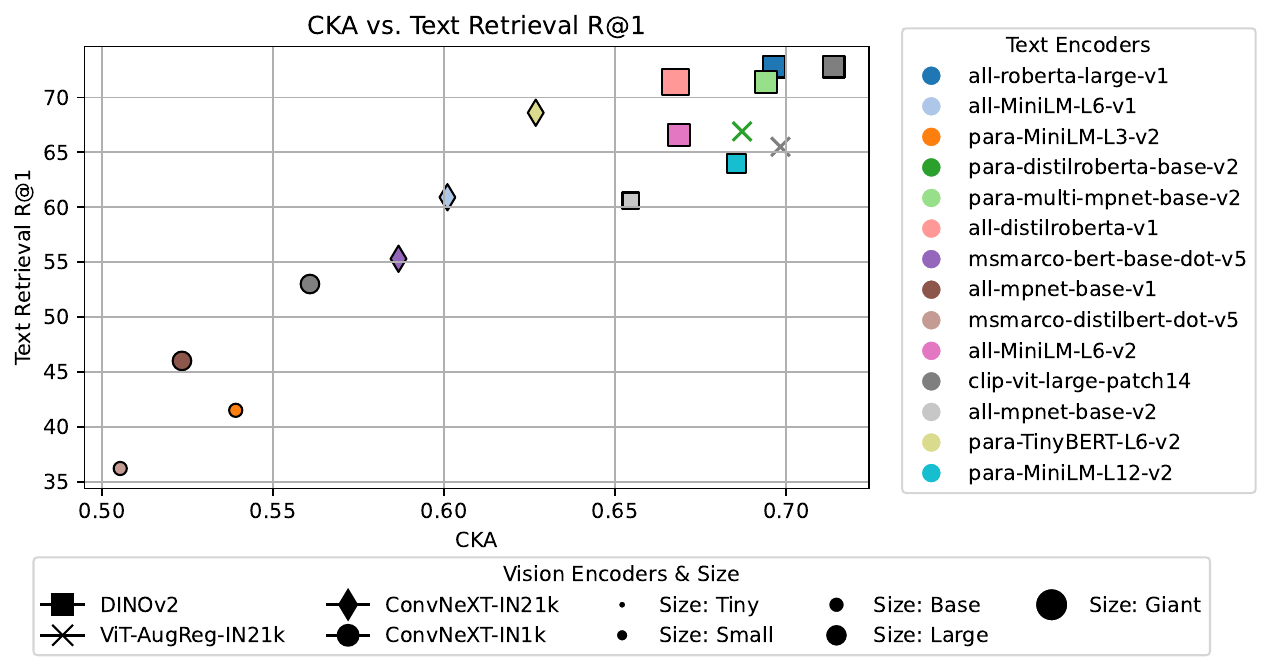}
        \vspace{-15pt}
        \setlength{\belowcaptionskip}{-12pt}
        \caption{\textbf{Retrieval performance vs. CKA for different encoder pairs.} Text retrieval accuracies on Flickr30k are compared to CKA, calculated on the COCO val set. Projectors are trained on the COCO train set. A clear correlation exists between CKA and alignment quality, as reflected in the retrieval accuracies.}
        \label{fig:cka_vs_ret}
    \end{minipage}%
    \hspace{0.02\linewidth}
    \begin{minipage}[c]{0.37\linewidth}
        \resizebox{\linewidth}{!}{%
            \begin{tabular}{lllll}
            \toprule
            V Proj. & V Proj. & T Proj. & T Proj. & INet \\
             Local &  CLS & Local & Global & 0-shot \\
            \midrule
            mlp                    & identity             & identity             & identity              & 68.81                     \\
            token                  & identity             & identity             & identity              & 68.84                            \\
            token                  & identity             & identity             & mlp                   & 70.90                            \\
            token                  & identity             & patch                & identity              & 71.85                            \\
            token                  & identity             & token                & mlp                   & 72.15                            \\
            identity                  & token                & token                & mlp                   & 75.53 \\
            token                  & token                & token                & mlp                   & \textbf{76.12}\\
            \bottomrule
            \end{tabular}
        }
        \setlength{\belowcaptionskip}{-10pt}
        \captionof{table}{\textbf{Projector ablations.}}
        \label{tab:ablation_proj}
    
        \vspace{12pt}
    
        \resizebox{\linewidth}{!}{%
            \begin{tabular}{lllll}
            \toprule
            Data Source           & N    & ImageNet & I2T   & T2I   \\
            \midrule
            LAION-CLASS-Collected & 6M   & 76.12         & 52.70 & 42.48 \\
            CC3M, CC12M, SBU      & 14M  & 54.17         & 85.30 & 72.44 \\
            Both                  & 20M  & 75.04         & 81.32 & 71.38 \\
            Both longer training  & 20M  & \textbf{76.30} & \textbf{87.54} & \textbf{74.17} \\
            \bottomrule
            \end{tabular}
        }
        \setlength{\belowcaptionskip}{-12pt}
        \captionof{table}{\textbf{Ablation of Alignment Training Data.}}
        \label{tab:ablation_data}
    \end{minipage}
    \setlength{\belowcaptionskip}{-15pt}
\end{figure*}

\section{Ablation Experiments}
\label{sec:ablation}

We present a set of ablations to validate different components of our pipeline empirically: CKA for encoder selection \ref{sec:cka_ablation}, the projector architecture and configuration \ref{sec:projector_ablation}, the alignment datasets, and the impact of class-collected data \ref{sec:dataset_ablation}. We evaluate on downstream tasks like 0-shot domain transfer to Imagenet classification and COCO / Flickr30k image-text retrieval scores.

\subsection{Effectiveness of CKA for encoder pair selection}
\label{sec:cka_ablation}

We train our projector configurations on various combinations of unimodal encoders using the COCO dataset and evaluate image/text retrieval accuracies on the Flickr30k test set, plotting these against CKA scores in Figure \ref{fig:cka_vs_ret}. The CKA, calculated on the COCO image-caption pairs, shows a strong correlation with retrieval accuracy, indicating that higher semantic similarity, as measured by CKA, predicts better alignment in image/text retrieval. Our findings suggest that CKA can effectively predict which encoder pairs will align well with projector training. The DINOv2-Large and CLIP-ViT-Large-text combination achieves the highest retrieval score, but certain unimodal pairs, like DINOv2-Large and All-Roberta-Large-v1 (CKA = 0.69), perform nearly as well. This indicates that these unimodal encoders are highly effective for vision-language alignment, leading us to choose the \textbf{DINOv2-Large} and \textbf{All-Roberta-Large-v1} pair for larger-scale experiments. Image Retrieval performance is illustrated in \ref{fig:cka_vs_retrieval_both}. Additionally, our findings indicate that CKA serves as a more reliable and straightforward metric for assessing alignment quality compared to other encoder pair selection strategies, such as downstream task performances, which tend to vary significantly depending on the specific task chosen(See Sec \ref{sec:encoder_pair_ablations_suppl}).



\subsection{Impact of Projector Architectures}
\label{sec:projector_ablation}

We ablate our projector combinations (\ref{tab:ablation_proj}) for the DINOv2 and All-Roberta-Large-v1 encoders by training the projectors to convergence on the LAION-Class-Collected dataset and evaluating the performance on ImageNet 0-shot domain transfer. An MLP applied solely to the local vision tokens achieved 68.81\% accuracy, while a Token projection \cite{Mukhoti_2023_CVPR} performed slightly better. Therefore, we used the Token projector for all tokens, both visual and textual.
Adding projectors to the text side, targeting both text tokens and a global projector on the averaged local tokens (rows 3, 4, and 5), resulted in performance improvements. These projectors help transform the unimodal text encoder's language-only representations to be more similar to the visual representations.
Introducing projectors to the CLS token (row 6) of the visual encoder led to a significant performance increase from 72.15\% to 75.13\%. Using both CLS and patch projectors in tandem yielded the best performance at 76.12\%. This improvement is attributed to DINOv2's dual training objectives: the image-level DINO \cite{caron2021emerging} objective on the CLS token and the patch-level iBOT \cite{zhou2021ibot} objective on the patch tokens learning effective global and local features. 

\subsection{Impact of Class-Collected Data / Retrieval Data}
\label{sec:dataset_ablation}

In Table \ref{tab:ablation_data}, we ablate the different components of our alignment data. Specifically, we compare the high concept coverage LAION-CLASS-Collected dataset with the high semantic alignment retrieval datasets: CC3M, CC12M, and SBU. Our experiments show that aligning DINOv2 and All-Roberta-Large-v1 on the high concept coverage dataset results in a high ImageNet zero-shot domain transfer accuracy of 76.1 \(\%\), though the retrieval accuracies are lower, at 52.7\(\%\)/42.2\(\%\) due to the noisy semantic alignment in LAION dataset. In contrast, training with the higher image-caption quality retrieval datasets results in high image and text retrieval scores on the Flickr30k val set (85.3\(\%\) and 72.4\(\%\), respectively). However, the limited concept coverage of these datasets leads to a lower ImageNet accuracy of 54.1\(\%\). Combining both types of datasets yields both high ImageNet accuracy and high image/text retrieval accuracies verifying that both dense coverage of unimodal spaces as well as high cross-modal semantic similarity is required to train effective projectors. To ensure that the extra data is adequately utilized, we train for an additional 15 epochs resulting in our best-performing model, achieving an ImageNet accuracy of 76.30\(\%\) and Flickr retrieval scores of 87.54\(\%\)/74.17\(\%\) (last row).

\section{Results}
We evaluate the alignment between vision and text encoders across commonly used VLM benchmarks, including zero-shot domain transfer, image retrieval \ref{sec:results-cls-ret}, localization \ref{sec:localization}, multilingual classification/retrieval \ref{subsec:multiling}, and dense caption image-text retrieval \ref{subsec:dci}. Our goal here is to evaluate the effectiveness of the learned alignment, showcase the flexibility of the framework as well as show that strong task-specific capabilities of uni-modal embeddings are retained in the joint embedding space.
We demonstrate that aligning unimodal vision-language encoders can match or exceed the performance of large CLIP models, despite using smaller datasets and less compute. Additionally, our alignment framework is flexible, enabling the use of specialized encoders for specific tasks, such as aligning multilingual text encoders for multilingual or low-resource image classification/retrieval, or long-context text encoders for dense image/caption retrieval. Furthermore, aligning DINOv2 with a text encoder improves image localization beyond CLIP's vision encoder due to DINOv2’s superior localization features.

\begin{table*}[]
\resizebox{\linewidth}{!}{%

\begin{tabular}{lllllllllllll}
\toprule
Model                  & N   & ImageNet & ImageNetv2 & Caltech & Pets & Cars & Flowers & Food & Aircrafts & SUN  & CUB  & UCF101 \\
\midrule
LAION-CLIP VIT-L       & 400M & 72.7     & 65.4       & 92.5    & 91.5 & \textbf{89.6} & 73.0    & \underline{90.0} & 24.6      & 70.9 & \textbf{71.4} & 71.6   \\
OpenAI-CLIP VIT-L      & 400M & 75.3     & \textbf{69.8}       & \underline{92.6}    & \textbf{93.5} & \underline{77.3} & \textbf{78.7}    & \textbf{92.9} & \textbf{36.1}      & 67.7 & 61.4 & \textbf{75.0}   \\
LiT L16L & 112M & \underline{75.7} & 66.6 & 89.1 & 83.3 & 24.3 & 76.3 & 81.1 & 15.2 & 62.5 & 58.7 & 60.0 \\
DINOv2-MpNet (Ours)    & 20M  & 74.8     & 68.0       & 91.8    & 91.7 & 71.0 & 75.8    & 87.5 & 23.0      & \underline{71.9} & 63.2 & 71.0   \\
DINOv2-ARL(Ours)       & 20M  & \textbf{76.3}     & \underline{69.2}       & \textbf{92.8}    & \underline{92.1} & 73.9 & \underline{78.4}    & 89.1 & \underline{28.1}      & \textbf{72.6} & \underline{66.1} & \underline{73.2}   \\
\bottomrule
\end{tabular}
}
\setlength{\belowcaptionskip}{-10pt}
\caption{\textbf{0-shot domain transfer to classification datasets.} We compare the performance of our DINOv2-ARL projector model, trained on a 20M dataset, against CLIP models from OpenAI and LAION across various datasets. Despite the smaller training size, our model achieves a 76.3\(\%\) accuracy on ImageNet, outperforming comparably sized CLIP models.} 
\label{tab:clas}
\end{table*}

\subsection{0-shot classification and Retrieval}
\label{sec:results-cls-ret}

\begin{table}[H]
\centering
\resizebox{0.8\linewidth}{!}{%
\begin{tabular}{lcccc}
\toprule
Model & \multicolumn{2}{c}{Flickr} & \multicolumn{2}{c}{COCO} \\
      & I2T & T2I & I2T & T2I \\
\midrule
LAION-CLIP VIT-L         & \textbf{87.6}       & 70.2       & 59.7     & 43.0     \\
OpenAI-CLIP VIT-L        & 85.2       & 64.9       & 56.3     & 36.5     \\
LiT L16L & 73.0 & 53.4 & 48.5 & 31.2 \\
DINOv2-MpNet (Ours)      & 84.6       & 71.2       & 58.0     & 42.6     \\
DINOv2-ARL (Ours)        & 87.5       & \textbf{74.1}       & \textbf{60.1}     & \textbf{45.1}     \\
\bottomrule
\end{tabular}
}
\setlength{\belowcaptionskip}{-15pt}
\vspace{-5pt}
\caption{\textbf{Image, Text Retrieval on COCO/Flickr30k.} Our Projector models show comparable text retrieval scores and significantly better image retrieval results.}
\label{tab:ret}
\end{table}
In this section we aim to evaluate the effectiveness of simple projection transformations in learning an alignment between semantically similar embedding spaces. Tables \ref{tab:clas} and \ref{tab:ret} report our model's performance on zero-shot domain transfer to image classification datasets and image-text retrieval on the Flickr30k and COCO datasets, respectively. Detailed descriptions of the evaluation datasets can be found in the \ref{sec:eval_datasets}, highlighting dataset domains, sizes, and prompt descriptions. We see that despite being trained on a 20M dataset our DINOv2-ARL projector model achieves an ImageNet accuracy of 76.3 \(\%\) which is 1 \(\%\) and 3.6 \(\%\) better than comparably sized CLIP models from OpenAI \cite{radford2021learning} and LAION \cite{schuhmann2021laion} respectively. Our DINOv2-ARL model demonstrates competitive performance across various datasets compared to LAION and OpenAI CLIP models . The relative performance of these models varies depending on the specific dataset. For example, on the Stanford Cars dataset, LAION-400m \cite{schuhmann2021laion} CLIP outperforms OpenAI CLIP by a significant margin of over 12\%. Conversely, for the Aircrafts dataset, both OpenAI CLIP and our DINOv2-ARL model show superior performance compared to LAION-400m CLIP. We believe this to be due to the differences in concept coverage for these particular datasets between the LAION400m, OpenAI WIT, and our MIX-CLASS-Collected datasets.

In text retrieval, our model outperforms or matches the next best CLIP model, LAION400M-CLIP VIT-L, with scores of 87.5\(\%\) vs 87.6\(\%\) on Flickr and 59.7\(\%\) vs 60.1\(\%\) on COCO. For image retrieval, our models show a significant advantage, achieving scores of 74.1\(\%\) vs 70.2\(\%\) on Flickr and 45.1\(\%\) vs 43.0\(\%\) on COCO. This improvement is likely due to the superior quality of the unimodal features produced by the DINOv2 and All-Roberta-Large-v1 encoders, compared to those of the multi-modal vision and text embeddings in the CLIP models. These results demonstrate that simple projector transformations between uni-modal encoders can achieve competitive performance similar to models trained from scratch, providing further evidence that simple projection transformations separate semantically similar embedding spaces.

\subsection{0-shot Localization}
\label{sec:localization}

One key advantage of leveraging frozen unimodal vision and text encoders is the enhancement provided by unimodal features. Specifically, the DINOv2 vision encoder's robust localization capabilities enhance the joint embedding space of the DINOv2-ARL model when trained solely with projectors. We assess this through zero-shot segmentation performance, similar to the \cite{bicaimproving, Mukhoti_2023_CVPR}, as shown in Table \ref{tab:localization}. Our approach involves computing cosine similarities between each patch and all the ground truth classes and subsequently upscaling similarity maps to the target size. Each patch is then classified into a corresponding class. Consistent with previous studies, the intersection over union (IoU) is computed solely for the foreground classes.(Refer to Sec. \ref{sec:seg-eval-details} for implementation details)

\begin{table}[ht]
\centering

\resizebox{0.7\linewidth}{!}
{%
\begin{tabular}{lll}
\toprule
Model        & Pascal & Pascal\\
             & VOC & Context \\
\midrule
OpenAI-CLIP-VIT-L* & 23.46      & 14.25          \\
SPARC              & 27.36      & 21.65          \\
DINOv2-ARL         & \textbf{31.37}      & \textbf{24.61}         \\
\bottomrule
\end{tabular}
}
\setlength{\belowcaptionskip}{-15pt}
\caption{\textbf{0-shot semantic segmentation mean IOU.} The table shows significant improvements by DINOv2-ARL, even without fine-grained alignment loss. * uses MaskCLIP trick.}
\label{tab:localization}
\end{table}

Our DINOv2-ARL model demonstrates superior performance compared to jointly trained dual encoder models like OpenAI's CLIP, achieving over 8\% improvement on Pascal VOC and over 10\% on Pascal Context. Notably, models utilizing a fine-grained alignment loss like SPARC \cite{bicaimproving} show improvements over CLIP. However, our DINOv2-ARL model outperforms SPARC by 4\% on VOC and 3\% on Context datasets. This underscores that the strong localization abilities of DINOv2 patch embeddings are retained even without training with a fine-grained alignment loss. We hypothesize that the localization performance could also benefit from a more precise localization alignment. Exploring fine-grained losses like SPARC with projector-only models presents an exciting direction for enhancing localization capabilities in VLMs. Additionally, the lower data demands of projector training may allow for the effective use of high-quality, smaller-scale grounding datasets to achieve precise alignment between word tokens and image patches in a supervised manner.

\subsection{Multi-Lingual Results}
\label{subsec:multiling}

\begin{table*}[ht]
\centering
\resizebox{\textwidth}{!}{%
\begin{tabular}{lrrrrrrrrrrrr}
\toprule
model & \multicolumn{6}{r}{classification} & \multicolumn{6}{r}{retrieval} \\
 & EN & DE & FR & JP & RU & average & EN & DE & FR & JP & RU & average \\
\midrule
nllb-clip-base@v1 & 25.4 & 23.3 & 23.9 & 21.7 & 23.0 & 23.5 & 47.2 & 43.3 & 45.0 & 37.9 & 40.6 & 42.8 \\
M-CLIP/XLM-Roberta-Large-Vit-B-32 & 46.2 & 43.3 & 43.3 & 31.6 & 38.8 & 40.6 & 48.5 & 46.9 & 46.1 & 35.0 & 43.2 & 43.9 \\
M-CLIP/XLM-Roberta-Large-Vit-L-14 & 54.7 & 51.9 & 51.6 & 37.2 & 47.4 & 48.6 & 56.3 & 52.2 & 51.8 & 41.5 & 48.4 & 50.0 \\
xlm-roberta-base-ViT-B-32@laion5b & 63.0 & 55.8 & 53.8 & 37.3 & 40.3 & 50.0 & 63.2 & 54.5 & 55.7 & 47.1 & 50.3 & 54.2 \\
nllb-clip-large@v1 & 39.1 & 36.2 & 36.0 & 32.0 & 33.9 & 35.4 & 59.9 & 56.5 & 56.0 & \textbf{49.3} & 50.4 & 54.4 \\
M-CLIP/XLM-Roberta-Large-Vit-B-16Plus & 48.0 & 46.1 & 45.4 & 32.9 & 40.3 & 42.5 & 63.2 & \textbf{61.4} & 59.3 & 48.3 & \textbf{54.8} & 57.4 \\
\midrule
ViT-L-14@laion400m & 72.3 & 48.2 & 49.9 & 2.7 & 4.5 & 35.5 & 64.5 & 26.7 & 38.3 & 1.4 & 1.7 & 26.5 \\
openai/clip-vit-large-patch14 & \textbf{75.6} & 46.7 & 49.6 & 6.6 & 3.5 & 36.4 & 59.4 & 19.9 & 28.5 & 4.1 & 1.3 & 22.6 \\
DINOv2-MpNet (Ours) & 73.4 & \textbf{61.6} & \textbf{58.3} & \textbf{43.2} & \textbf{49.3} & \textbf{57.1} & \textbf{70.7} & 60.6 & \textbf{60.6} & 45.6 & 52.7 & \textbf{58.0} \\
\bottomrule
\end{tabular}
}
\vspace{-5pt}
\caption{\textbf{Multilingual Classification and Image-Caption Retrieval.} Performance comparison of DINOv2-MpNet with various CLIP models and multilingual baselines on multilingual ImageNet and XTD datasets. Despite being trained only on English data, DINOv2-MpNet outperforms models trained on multiple languages. The upper half of the tables shows multilingual-trained models, while the lower half lists models trained only on English data.}
\label{tab:multi_combined_table}
\vspace{-10pt}
\end{table*}

Similar to the previous section, here we assess whether multi-lingual capabilities of a language encoder is retained when aligned to a vision encoder using projectors. 
We demonstrate this by aligning DINOv2-Large with paraphrase-multilingual-MpNetv2 (referred to as MpNet), chosen for its high CKA compatibility, using only English image-caption pairs and  evaluating model performance on multi-lingual image retrieval on the XTD dataset \cite{aggarwal2020zeroshot} and classification on the ImageNet dataset. For classification, we translated the prompts to the considered languages using nllb-200-distilled-600M \cite{costa2022no}.
Multi-lingual classification and retrieval results for five representative languages are presented in Table \ref{tab:multi_combined_table} (For Detailed results Refer to Sec \ref{sec:multi_full} ) .The lower section lists models trained exclusively with English captions, \cite{radford2021learning} \cite{schuhmann2021laion} while upper sections feature models trained with multi-lingual captions \cite{schuhmann2022laion},\cite{chen2023mclip}, \cite{visheratin2023nllb}. 

Our DINOv2-MpNet, trained solely on English image-caption pairs, outperforms other English-only CLIP models by over 31\% in average retrieval performance across five languages and by 6\% in English. DINOv2-MpNet remains competitive across both Latin and non-Latin languages, even against models trained on multilingual data. Notably, it outperforms the LAION5B trained xlm-roberta-base-VitB32 by 0.6\%, despite using only 20 million English image-caption pairs compared to over 2B non-English pairs in LAION5B. A similar trend is observed in classification, with DINOv2-MpNet surpassing the next best English-trained model, by over 20\% on average across five languages. Among multilingual models, the next best M-CLIP/XLM-Roberta-Large-Vit-L-14 by over 8\%, despite not using any multilingual text data. DINOv2-MpNet’s robust multilingual performance, achieved without multilingual training data, demonstrates that MpNet's capabilities are preserved in the joint embedding space through effective projector training of unimodal models.

\subsection{Densely Captioned Images (DCI) Dataset and Long-Text Retrieval}
\label{subsec:dci}

We assess whether the ARL model maintains its long-context capabilities in the joint embedding space by conducting image and long caption retrieval on the Densely Captioned Images (DCI) dataset \cite{urbanek2024picture}, which features caption pairs averaging over 1,000 words. Unlike DCI’s benchmarks that use summarized captions (see \ref{sec:sdci}), we focus on full image-text and text-image retrieval tasks without summarization or subcropping, enabling a comprehensive evaluation of our framework’s long-text retrieval capabilities.

\begin{figure}
    \centering
    \includegraphics[width=\linewidth]{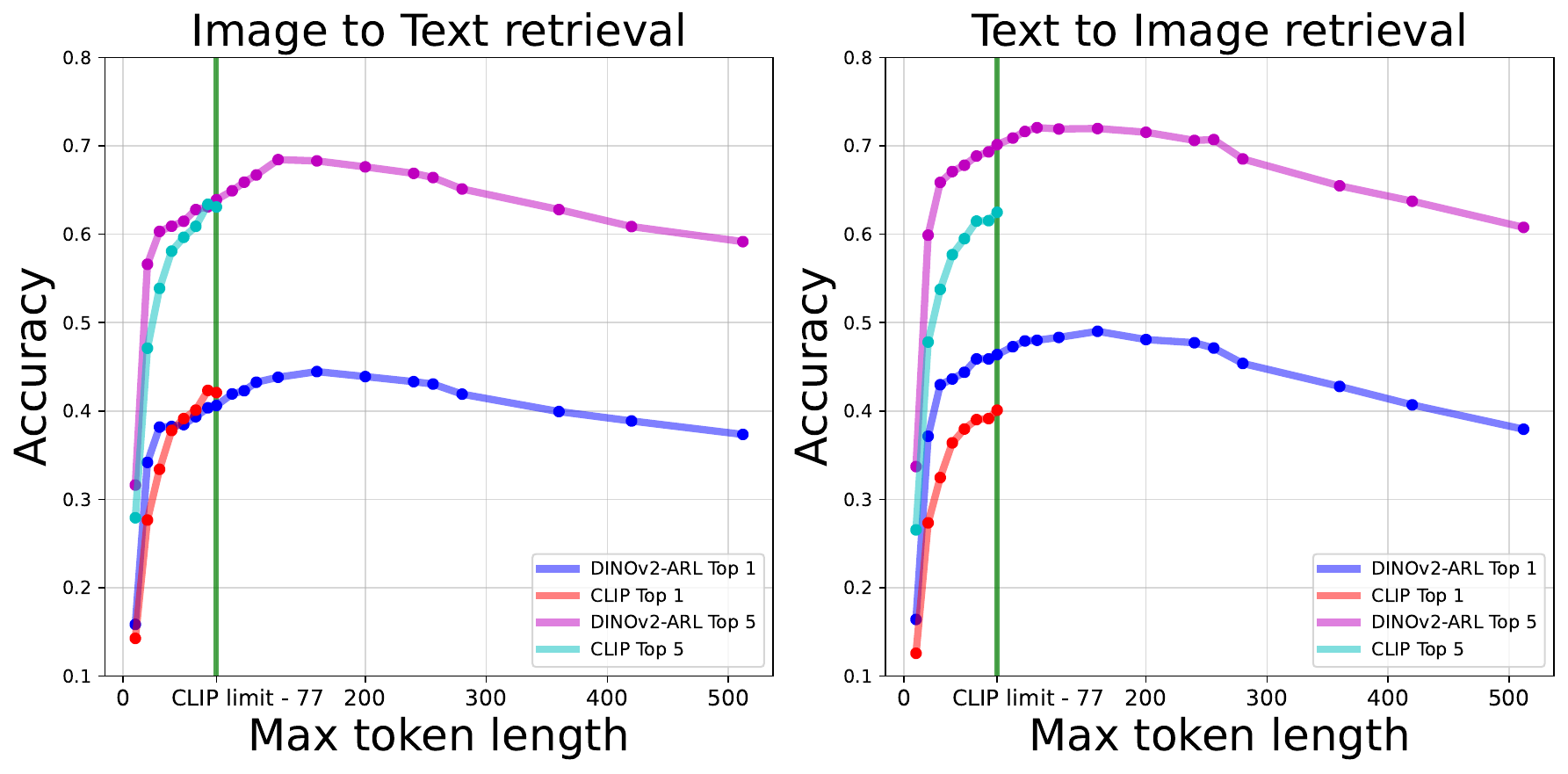}
    \vspace{-15pt}
    \setlength{\belowcaptionskip}{-12pt}
    \caption{Retrieval performance comparison between DINOv2-ARL encoder pair and OpenAI CLIP as the maximum token length increases. The vertical green line indicates the standard CLIP token limit of 77.}
    \label{fig:dci_figure}
\end{figure}

To demonstrate the retention of long-context ability, we conducted an experiment varying the maximum token length allowed by the tokenizer. As shown in Figure~\ref{fig:dci_figure}, our DINOv2-ARL encoder pair achieves comparable performance to OpenAI CLIP at the standard limit of 77 tokens. However, our approach's strength becomes evident as we extend beyond this limit, with consistent improvement in retrieval accuracy up to approximately 200-300 tokens. Given that DINOv2-ARL was trained with short-context image-caption pairs, these results underscore the model’s ability to retain long-context capabilities in the aligned joint embedding space.

\subsection{Alignment Compute}
\label{subsec:compute}

We report the Alignment Training compute requirements for different models in \ref{tab:model_compute_comparison}. We see that aligning pre-trained vision, language encoders to get a competitive CLIP like model requires only 50 hours of training with 8 A100 GPUS which is almost a 65 fold reduction in the amount of alignment compute. This makes the development of multi-modal models accessible to the wider research community as well as reducing the environmental impact of training highly performant multi-modal models by reusing strong publicly available uni-modal models.  Since we only need to train 11.5M of the total 670M parameters (about 1 \(\%\)) we can train with a much smaller and denser dataset reducing the data requirements to 20M which is 20 fold decrease in dataset requirement compared to CLIP models from LAION and OpenAI making our framework useful for training performant multi-modal models in various domains like multi-modal systems for low-resource languages, 3D model search systems, fMRI to Image model mapping systems and many more where paired data is limited. Despite the reduced compute and data requirements for alignment, our model outperforms both CLIP models compared on domain transfer to Imagenet as well as image, text retrieval. Caveat: Our alignment assumes strong unimodal encoders are available and does not account for training compute. For completeness, DINOv2 was trained with 22k GPU (A100) hours, while ARL and MpNet used 7 TPUv3-8 for 400k steps \cite{reimers2024sentence-transformers}.

\begin{table}
    \centering
    \resizebox{\linewidth}{!}{%
    \begin{tabular}{lrrrrr}
    \hline
    \textbf{Model}               & \textbf{Data} & \textbf{SS} & \textbf{Trainable / Total} & \textbf{Compute}& \textbf{IN 0-shot} \\ \hline
    OpenAI CLIP        & 400M                     & 12.8B                      & 427M / 427M                           & 21,845                         & 72.7\%                    \\
    LAION400M CLIP      & 400M                     & 12.8B                      & 427M / 427M                           & 25,400                         & 75.3\%                    \\
    DINOv2-ARL       & 20M                      & 0.6B                       & 11.5M / 670M                          & 400                            & 76.3\%                    \\ \hline
    \end{tabular}%
    }
    \setlength{\belowcaptionskip}{-12pt}
    \captionof{table}{\textbf{Compute requirements, Dataset size, and Number of trainable parameters are orders of magnitude lower when using projectors to align semantically similar encoders.} By using projectors to align semantically similar encoders, compute requirements (for alignment) drop 65-fold, paired dataset size shrinks by 20 times, and only 1\% of total parameters are trainable while outperforming other CLIP models. Compute measured in GPU hours on an A100 (80 GB) GPU.}
    \label{tab:model_compute_comparison}
\end{table}

\section{Conclusion}
Our research presents a significant advancement in vision-language alignment, showing that high performance can be attained with considerably fewer resources than usually needed. By utilizing the inherent compatibility of well-trained unimodal encoders, we offer a new perspective on efficient multimodal AI development.

Future efforts could investigate how our models might be integrated with Large Language Models, employ fine-grained alignment techniques, utilize different projection architectures, and extend to additional modalities beyond vision and language. Our framework may facilitate more accessible multimodal AI research, potentially speeding up innovation and influencing future approaches to multimodal AI development.

{\small
\bibliographystyle{ieee_fullname}
\bibliography{egbib}
}

\newpage
\appendix

\setcounter{table}{0}
\setcounter{figure}{0}
\renewcommand{\thetable}{A.\arabic{table}}
\renewcommand{\thefigure}{A.\arabic{figure}}
\section{Appendix}

\begin{figure}
    \vspace{-0.6cm}
    \centering
    \includegraphics[width=0.75\columnwidth, trim={0.5cm 0.4cm 0.7cm 0.5cm},clip=true]{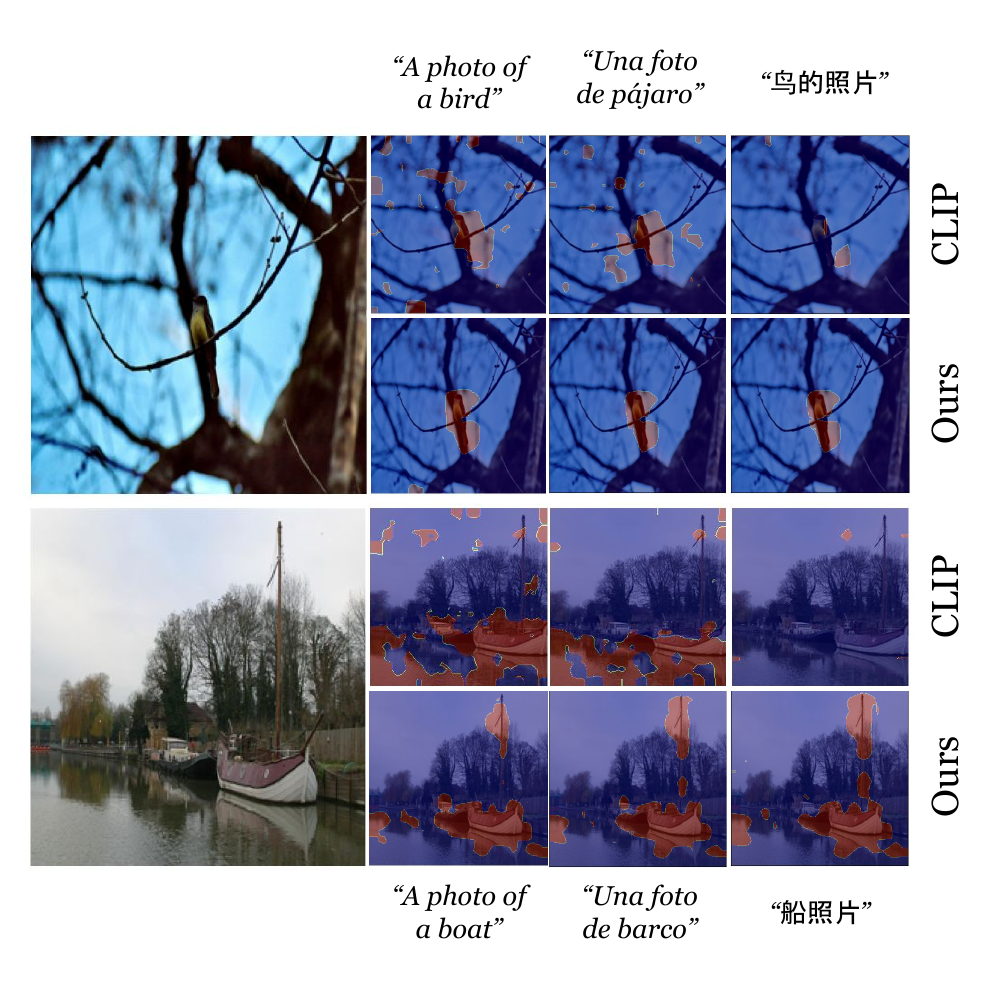}
    \vspace{-0.45cm}
    \caption{Compared to CLIP, our approach of aligning DINOv2-MpNet achieves improved segmentation maps focusing on the relevant objects in the multilingual setting.}
    \label{fig:multi-lingual-semantic-seg}
\end{figure}

\subsection{Unlocking parts of text and vision encoders}

We evaluated our model with different parts of the vision and text encoders unlocked for DinoV2-ARL, shown in Tab.\ref{tab:unlock}. Similar to Lit \cite{zhai2022lit} we find that unlocking the vision encoder (\eg, via BitFit \cite{zaken2021bitfit}) reduced performance, while full unlock resulted in unstable training. In contrast, unlocking the text encoder or applying $\text{BitFit}_{text}$ 
slightly improved performance with increased training costs.

\subsection{Training CLIP with same dataset}
We compare our approach against CLIP-ViT-L models trained from scratch, and projector-only trained in Tab. \ref{tab:clip}. We see that our 20M dataset is not enough to train the CLIP model (427M params) from scratch. Meanwhile, projector-only training of CLIP improves over OpenAI CLIP on COCO I2T and achieves competitive performance on Imagenet. Notably, none of the trained CLIP models outperform DINOv2-ARL.

\subsection{Multi-lingual 0-shot Semantic Segmentation}
The lower compute and paired data requirements of the framework lead to application flexibility simply by swapping the unimodal encoders. (see Sec. 6.2-6.4 in the main paper). An additional advantage of this flexibility is showcased in Fig. \ref{fig:multi-lingual-semantic-seg} and Tab. \ref{tab:multi-seg-ious}, where we use our aligned DINOv2-MpNet to perform multi-lingual semantic segmentation. Our segmentation scores stay consistent with different languages while CLIP often fails on non-english languages.

\begin{table}[h]
    \centering
    \begin{minipage}[L]{0.45\columnwidth}
        \centering
        \caption{Multilingual Segmentation IOU scores.\vspace{-0.3cm}}
        
        \resizebox{\columnwidth}{!}{
        \begin{tabular}{lrr}
        \hline
        Language &  CLIP &  \textit{DINOv2-MpNet} \\
        \hline
        EN & 23.46 & 29.07 \\
        ES & 18.86 & 28.69 \\
        ZH & 8.46 & 28.06 \\
        FR & 15.12 & 28.48 \\
        DE & 21.30 & 27.91 \\
        RU & 5.72 & 26.85 \\
        \hline
        \end{tabular}
        }
        \label{tab:multi-seg-ious}
    \end{minipage}
    \begin{minipage}[R]{0.53\columnwidth}
        \centering
        \vspace{-0.1cm}
        \caption{Unlocking Encoders.\vspace{-0.35cm}}
        
        \resizebox{\columnwidth}{!}{
        \begin{tabular}{lrr}
        \hline
        Method (15 epochs) &  Imagenet &  COCO I2T \\
        \hline
        $BitFit_{all}$ & 67.67 & 53.16\\
        $BitFit_{text}$ & 74.58 & 56.72\\
        Text unlock & \textbf{75.90} & \textbf{56.62}\\
        \textbf{Projectors}  & 75.04 & 56.32\\
        \hline
        \end{tabular}
        }
        \label{tab:unlock}
        \caption{CLIP on our dataset.}
        \vspace{-0.35cm}
        \resizebox{\columnwidth}{!}{
        \begin{tabular}{lrr}
        \hline
        Method (30 epochs)&  Imagenet &  COCO I2T \\
        \hline
        $CLIP_{scratch}$ & 50.30 & 36.12\\
        $CLIP_{openai}$ & 75.32 & 56.31\\
        $CLIP_{projectors}$ & 72.10 & 59.04 \\
        \textit{DINOv2-ARL} & \textbf{76.45} & \textbf{60.14}\\
        \hline
        \end{tabular}
        }
        \label{tab:clip}
    \end{minipage}
\end{table}

\subsection{Toy Example using Random Latent Model}
\label{sec:toy_example_synthetic}
Similar to Sec. \ref{sec:cka_toy_example} here we investigate whether semantically similar encoder embedding spaces can be aligned through a simple projection transformation, using a random latent model. 

\begin{figure}[h!]
    \centering
    \begin{minipage}[b]{0.45\textwidth}
        \centering
        \includegraphics[width=\linewidth]{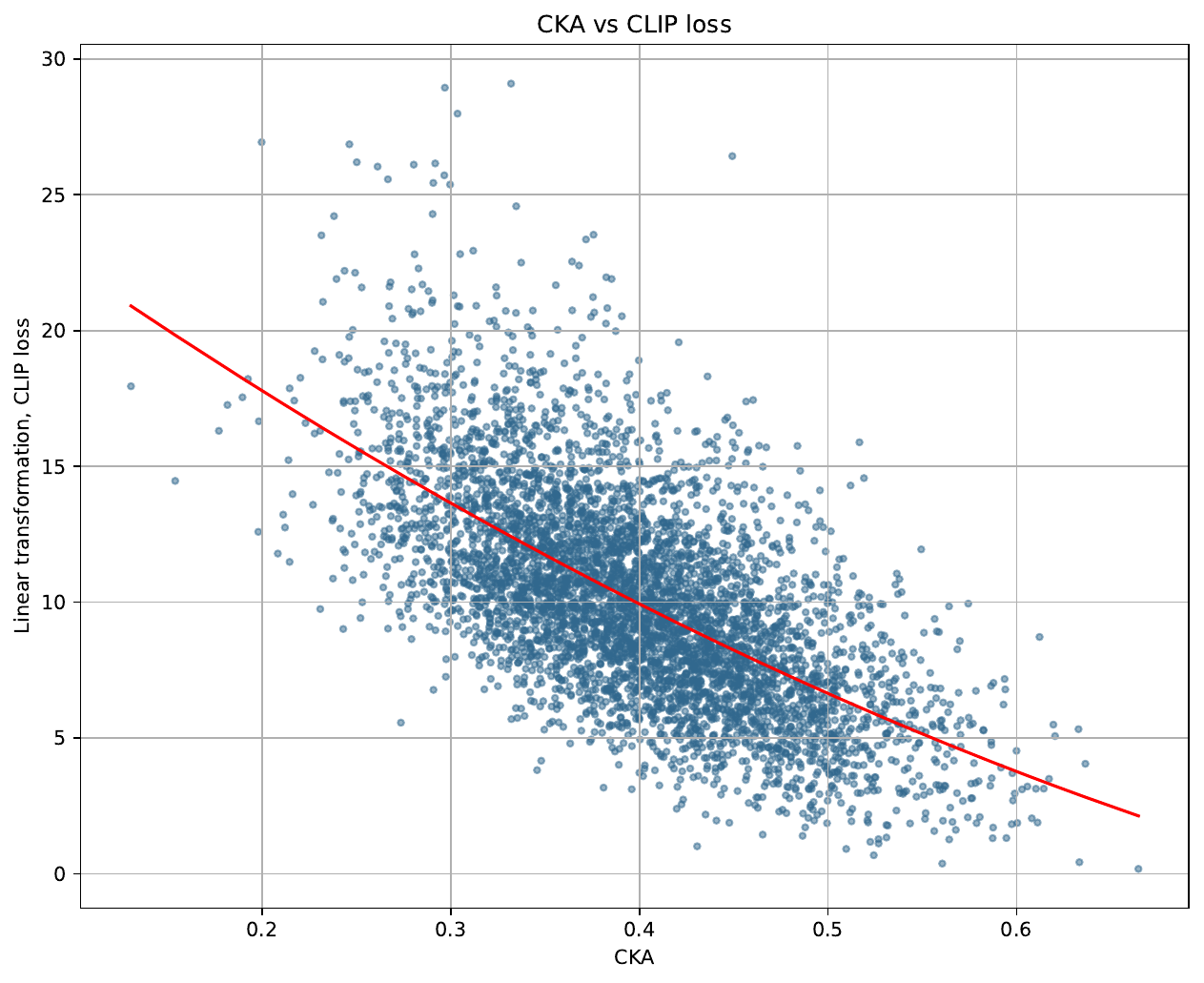}
        \caption{\textbf{CLIP Loss minima are negatively correlated to CKA.} We plot CKA vs CLIP Loss for random instances of A and B.}
        \label{fig:cka_vs_clip_random}
    \end{minipage}
    \hfill
    \begin{minipage}[b]{0.45\textwidth}
        \centering
        \begin{lstlisting}[language=Python, frame=single, xleftmargin=0.2cm, xrightmargin=0.2cm]
# Init Z with random values scaled to [-1, 1]
Z = 2 * rand(n, d) - 1

# Define non-linear transforms T1 and T2
T1, T2 = NLTransform(d, d), NLTransform(d, d)

# Sample random weights w1 and w2
w1, w2 = rand(1), rand(1)

# Compute A and B using transforms
A = T1(Z) + w1 * rand(n, d)
B = T2(Z) + w2 * rand(n, d)
        \end{lstlisting}
        \caption{\textbf{Code for initializing A and B from a latent world model Z.} Random instances of A, B are generated using random non-linear transformations of latent vector Z denoting a representation of the real world.}
        \label{fig:code}
    \end{minipage}
\end{figure}

In our experiment, we generated $10^{3}$ instances of two vector sets, $A$ and $B$, each containing 32 vectors of 16 dimensions. Following the approach in \cite{Maniparambil_2024_CVPR, huh2024platonic}, we modeled the world using a latent distribution $Z$, with Image and Text representations ($A$ and $B$) as random independent non-linear transformations from $Z$ with additive noise. For each sampled pair of $A$ and $B$ matrices, we calculated the CKA and the minimum CLIP loss. The non-linear transform was defined as a randomly initialized 2-layer MLP with ReLU non-linearity and hidden dimensions significantly larger than the input dimensions, ensuring it could universally approximate the non-linear transformation \cite{hornik1989multilayer}. Figure \ref{fig:code} was used to generate each instance.

Figure \ref{fig:cka_vs_clip_random} illustrates the results of this experiment, showing a clear negative correlation between CKA and minima of the CLIP loss. As CKA increases, indicating greater similarity between the similarity structures of A and B, the minima of CLIP loss consistently decreases. Despite arising from a simplified experiment, the observed strong inverse relationship between CKA and CLIP loss provides empirical support for using CKA as a predictor of alignment potential between embedding spaces. Since CLIP loss is lower-bounded by mutual information, and mutual information is correlated with HSIC, higher CKA suggests a stronger alignment between embeddings. This implies that the achievable minima of CLIP loss is lower when the embedding spaces already have a higher CKA, reflecting greater mutual information and ease of alignment.

\subsection{Embedding Graph structures visualized }

\label{sec:graph_structures}

\begin{figure}
    \centering
        \includegraphics[height=0.6\linewidth]{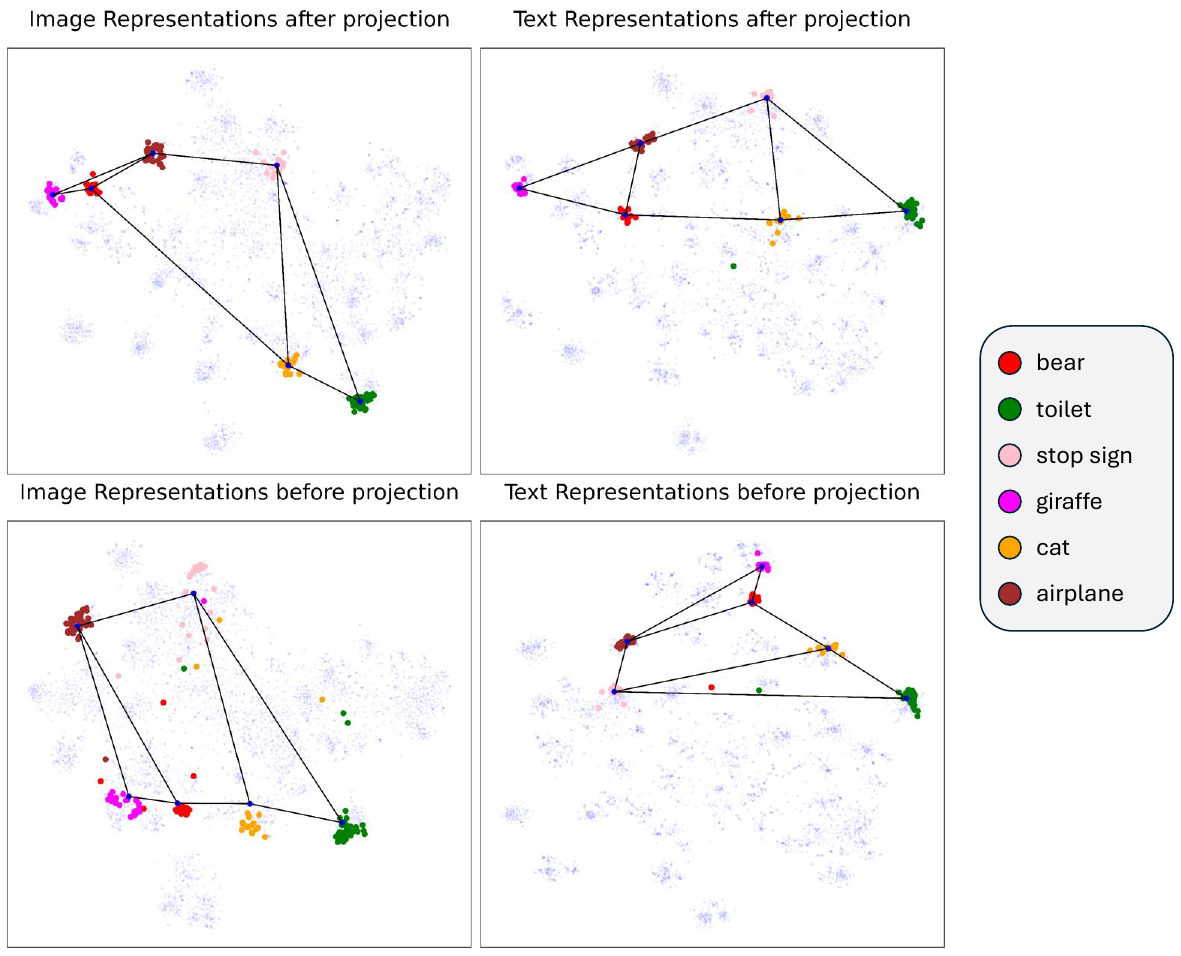}
        \includegraphics[height=0.6\linewidth]{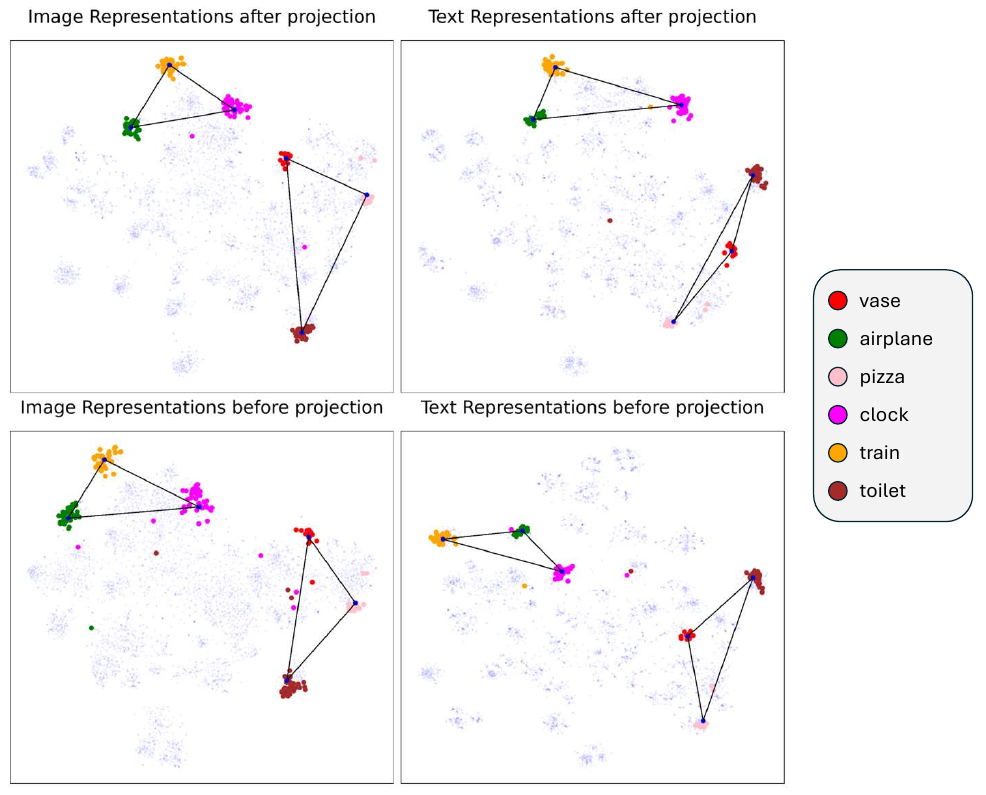}
        \includegraphics[height=0.6\linewidth]{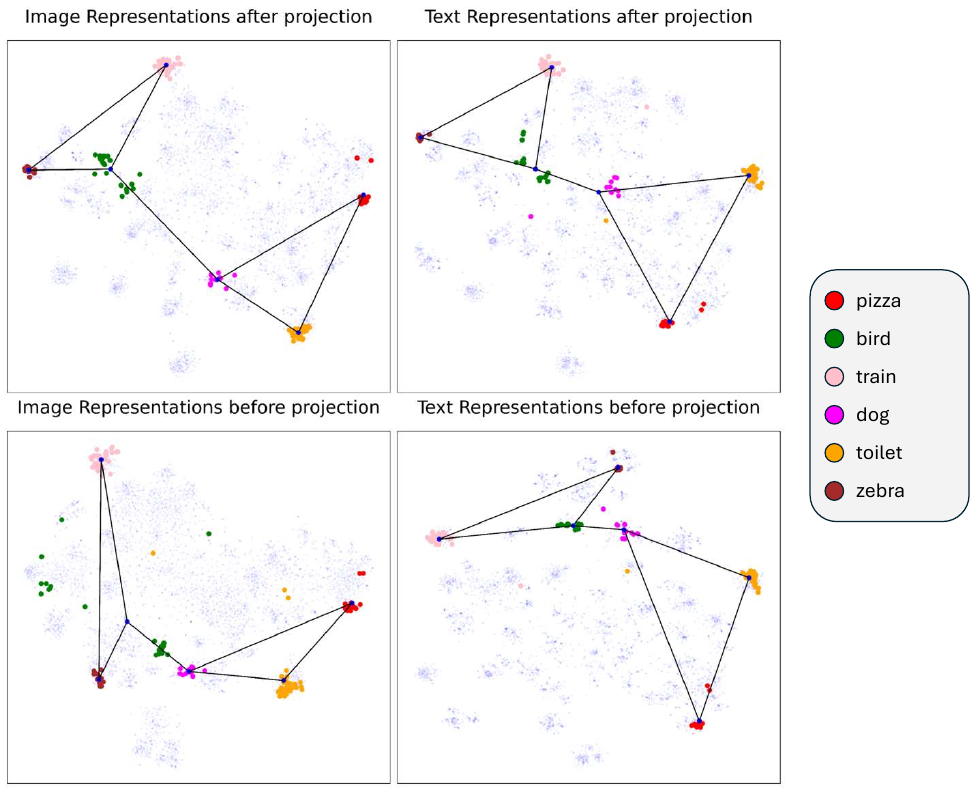}
        \caption{TSNE visualizations of encoder outputs for six COCO detection classes. Left: DINOv2 (vision), Right: All-Roberta-Large-v1 (text).}
        \label{fig:tsne_plot_1}
\end{figure}

To visually demonstrate how CKA represents similarities in graph structures across different encoder spaces, we conducted an experiment using the MSCOCO validation set. We examined encoder outputs for DINOv2 and All-Roberta-Large-v1, before and after projection, focusing on relationships between formed clusters in both domains. For each cluster, we identify COCO detection class and COCO image-caption pairs where the image contained only the respective class among its detection annotations. We then extracted encoder outputs for these samples from both vision and text encoders, before and after applying our projection layers, and applied the TSNE algorithm to visualize their structure in a lower-dimensional space. For each visualization, we pick 6 classes to highlight the shape similarities between graphs of encoder spaces.

Figure \ref{fig:tsne_plot_1} shows the resulting TSNE visualizations for the six selected classes across four conditions: vision pre-projection, vision post-projection, text pre-projection, and text post-projection. The visualizations reveal striking similarities in cluster shapes and relative positions across the different encoder spaces, particularly before projection. This visual similarity aligns with our quantitative CKA results, providing an intuitive illustration of how CKA captures structural similarities between different embedding spaces.

\subsection{Comparison to LiLT}
Tables \ref{tab:clas_with_lilt} and \ref{tab:ret_with_lilt} report the zero-shot domain classification and retrieval performance of LiLT models \cite{khancontrastive}. The vision encoder is initialized with the DeiT base model \cite{touvron2021training}, and the text encoder is from SimCSE \cite{gao2021simcse}. The LilT\textsubscript{DA}-base model is trained by duplicating and appending the last transformer layer, while only unlocking the last encoder and projector layers. The LilT\textsubscript{LwA}-base model introduces trainable layerwise adapters for both the vision and text encoders. LiLT public checkpoints are trained on 500k image-caption pairs from the COCO dataset. However, LiLT’s performance lags behind CLIP models and our DINOv2-ARL projector model, primarily due to suboptimal encoder pairs and limited concept coverage in the COCO training set for alignment.
\begin{table}
\resizebox{\linewidth}{!}{%

\begin{tabular}{lllllllllllll}
\toprule
Model                  & N   & ImageNet & ImageNetv2 & Caltech & Pets & Cars & Flowers & Food & Aircrafts & SUN  & CUB  & UCF101 \\
\midrule
LAION-CLIP VIT-L       & 400M & 72.7     & 65.4       & 92.5    & 91.5 & \textbf{89.6} & 73.0    & \underline{90.0} & 24.6      & 70.9 & \textbf{71.4} & 71.6   \\
OpenAI-CLIP VIT-L      & 400M & 75.3     & \textbf{69.8}       & \underline{92.6}    & \textbf{93.5} & \underline{77.3} & \textbf{78.7}    & \textbf{92.9} & \textbf{36.1}      & 67.7 & 61.4 & \textbf{75.0}   \\
LiT L16L & 112M & \underline{75.7} & 66.6 & 89.1 & 83.3 & 24.3 & 76.3 & 81.1 & 15.2 & 62.5 & 58.7 & 60.0 \\
LilT$_{DA}$-base & 0.5M & 15.9 & 12.9 & 37.6 & 7.2 & 1.6 & 1.1 & 13.3 & 1.7 & 25.6 & 2.3 & 19.1\\
LilT$_{LwA}$-base & 0.5M & 14.4 & 12.1 & 42.3 & 4.8 & 1.3 & 2.1 & 12.3 & 1.6 & 26.5 & 1.4 & 26.6\\
DINOv2-MpNet (Ours)    & 20M  & 74.8     & 68.0       & 91.8    & 91.7 & 71.0 & 75.8    & 87.5 & 23.0      & \underline{71.9} & 63.2 & 71.0   \\
DINOv2-ARL(Ours)       & 20M  & \textbf{76.3}     & \underline{69.2}       & \textbf{92.8}    & \underline{92.1} & 73.9 & \underline{78.4}    & 89.1 & \underline{28.1}      & \textbf{72.6} & \underline{66.1} & \underline{73.2}   \\
\bottomrule
\end{tabular}
}
\setlength{\belowcaptionskip}{-10pt}
\caption{\textbf{0-shot domain transfer to classification datasets.} We compare the performance of our DINOv2-ARL projector model, trained on a 20M dataset, against CLIP models from OpenAI and LAION across various datasets. Despite the smaller training size, our model achieves a 76.3\(\%\) accuracy on ImageNet, outperforming comparably sized CLIP models.} 
\label{tab:clas_with_lilt}
\end{table}

\begin{table}
\resizebox{\linewidth}{!}{%
\begin{tabular}{lcccc}
\toprule
Model & \multicolumn{2}{c}{Flickr} & \multicolumn{2}{c}{COCO} \\
      & I2T & T2I & I2T & T2I \\
\midrule
LAION-CLIP VIT-L         & \textbf{87.6}       & 70.2       & 59.7     & 43.0     \\
OpenAI-CLIP VIT-L        & 85.2       & 64.9       & 56.3     & 36.5     \\
LiT L16L                 & 73.0       & 53.4       & 48.5     & 31.2 \\
LilT$_{DA}$-base           & 47.6       & 34.46      & 41.4     & 29.1 \\
LilT$_{LwA}$-base          & 56.8       & 41.7       & 47.0     & 33.7 \\
DINOv2-MpNet (Ours)      & 84.6       & 71.2       & 58.0     & 42.6     \\
DINOv2-ARL (Ours)        & 87.5       & \textbf{74.1}       & \textbf{60.1}     & \textbf{45.1}     \\
\bottomrule
\end{tabular}
}
\caption{\textbf{Image, Text Retrieval on COCO/Flickr30k.} Our model shows comparable text retrieval scores and significantly better image retrieval results.}
\label{tab:ret_with_lilt}
\end{table}

\subsection{Encoder Pairs Ablations}
\label{sec:encoder_pair_ablations_suppl}
\begin{figure}[t]
    \centering
    \includegraphics[width=\linewidth]{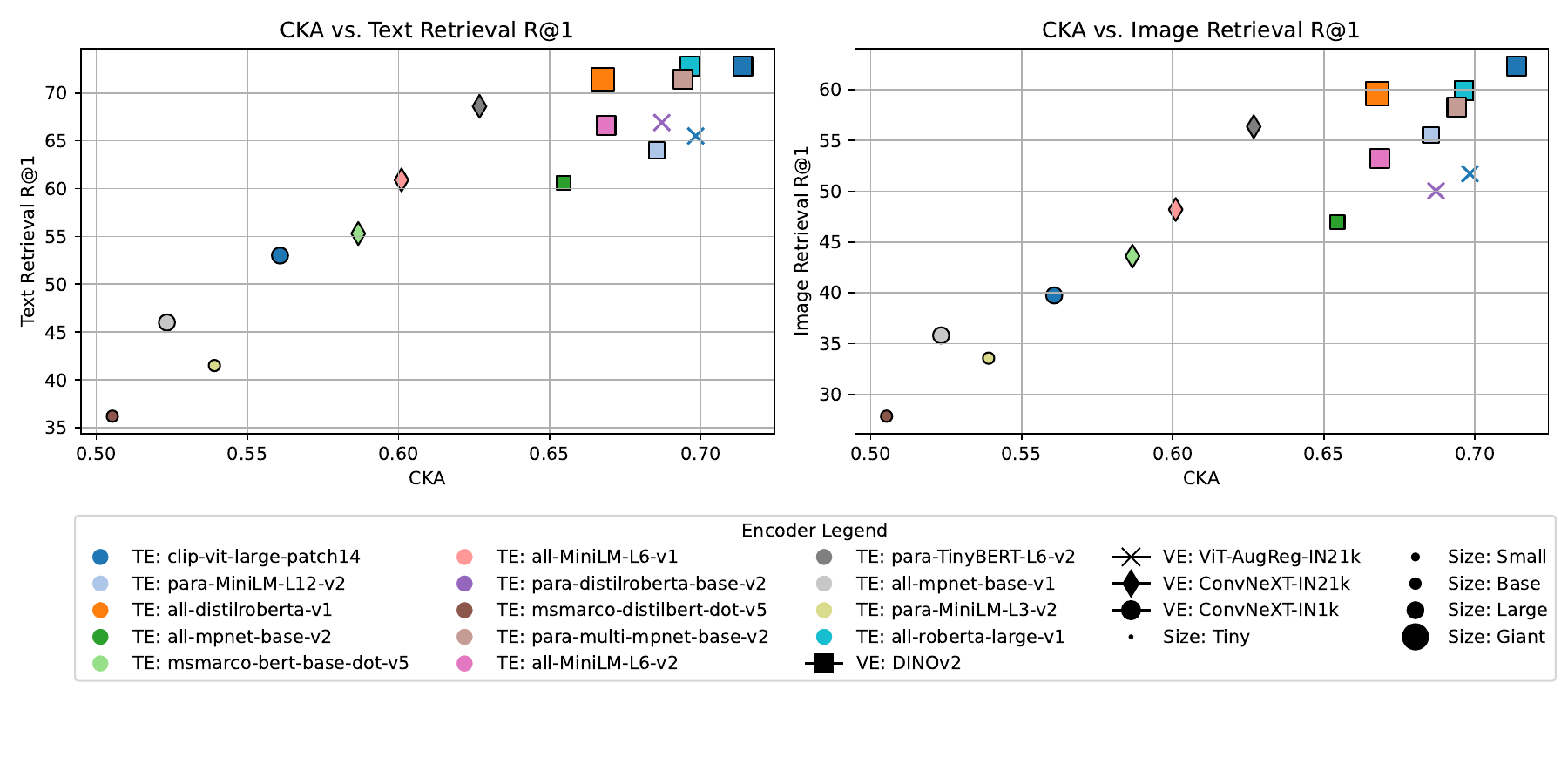}
    \caption{\textbf{Retrieval performance vs. CKA for different encoder pairs.} Text/Image retrieval accuracies on Flickr30k are compared to CKA, calculated on the COCO val set. Models trained on COCO train set. A clear correlation exists between CKA and alignment quality (Pearson correlation = 0.92, p = 2.1e-7), as reflected in retrieval accuracies.}
    \label{fig:cka_vs_retrieval_both}
\end{figure}

Similar to Sec \ref{sec:cka_ablation}, we train our projector configurations on various combinations of unimodal encoders using the COCO dataset and evaluate image/text retrieval accuracies on the Flickr30k
test set, plotting these against CKA scores. In Fig. \ref{fig:cka_vs_retrieval_both} both the Image and Text retrieval accuracies shows a strong correlation with CKA suggesting that CKA can effectively predict which encoder pairs will align well with projector training.

A naive approach to choosing the best encoder pair is to chose the unimodal encoders with highest performance in their respective modalities, but it's not straightforward which benchmarks can be more predictive of ease of alignment. To demonstrate this, we consider the same ablation as above, but with DINOv2 and 14 different text encoders from the SentenceTransformers \cite{sentencetransformer}
library. We consider 2 types of text model benchmarks.
1. Sentence Embedding task or Semantic Textual Similarity (STS) is the task of evaluating how similar two texts are in terms of meaning. These models take a source sentence and a list of sentences and return a list of similarity scores. The task is evaluated using Spearman’s Rank Correlation. We average over 14 datasets reported in \cite{sentencetransformer, reimers2024sentence-transformers}. 
2. Semantic Search (SS) is the task of retrieving relevant documents or passages based on the semantic content of a query. Rather than relying solely on keyword matching, semantic search models generate embeddings for both the query and the documents, allowing for retrieval based on contextual and conceptual similarity and is evaluated using Normalized Discounted Cumulative Gain (nDCG), which measure the relevance of retrieved documents in ranked lists. We average over 6 datasets reported in \cite{sentencetransformer, reimers2024sentence-transformers}. 

\begin{figure}[t]
    \centering
    \includegraphics[width=\linewidth]{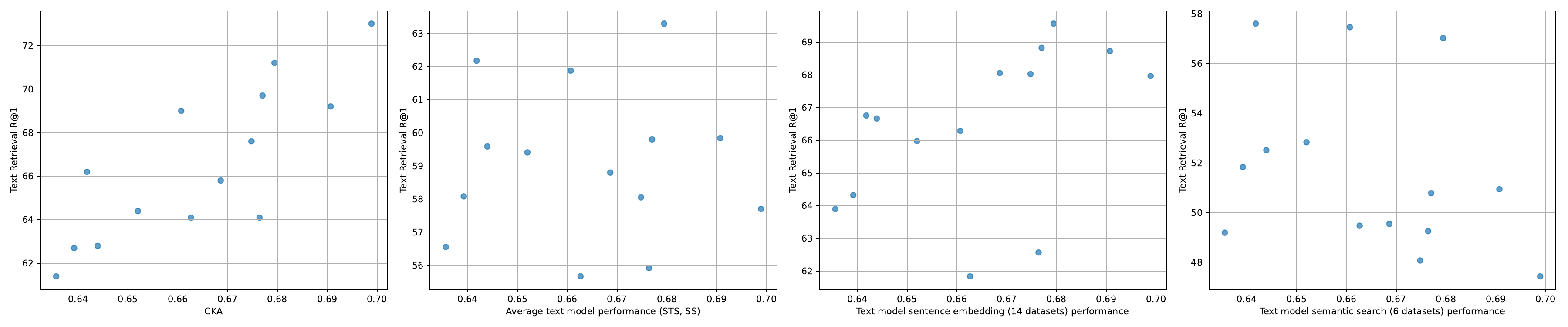}
    \caption{\textbf{Retrieval performance vs. text model performance for DINOv2 and different text encoders.} Text/Image retrieval accuracies on Flickr30k are compared different text encoder tasks performance. CKA is more closely correlated with retrieval performance than text encoder downstream task performance on sentence embedding tasks, semantic search tasks. 
    Models trained on COCO train set. }
    \label{fig:cka_vs_perf_all}
\end{figure}

In Fig \ref{fig:cka_vs_perf_all}, we see that there is a clear correlation (pearson corr.=0.81, p=4e-4) between downstream Flickr30k performance and CKA on the COCO val set, suggesting that CKA is a better predictor of ease of alignment. The average unimodal performance (pearson corr.=0.47, p=0.08), as well as the semantic search (SS) performance (pearson corr.=0.13, p=0.65), are not predictive of the ease of alignment. Meanwhile, Sentence Task Similarity (STS) tasks are more predictive of downstream alignment (pearson corr.=0.72, p=0.003) but still worse than CKA and it's not intuitive which unimodal performance is to be considered.

\subsection{Data Curation Implementation Details}
\label{sec:curation_impl}
We streamline our class collection process by precomputing CLIP text embeddings for LAION-400M and CLIP image prototype embeddings for various concepts, allowing us to run different collection methods without needing to recompute embeddings. The embedding process takes just 12 hours on two nodes with 4 A6000 GPUs each. Class-level collection is performed using GPU-accelerated PyTorch code on a single GPU, completing in under an hour. While image-to-image-prototype collection, as in \cite{oquab2023dinov2}, could yield higher-quality results, it demands significantly more GPU resources due to the need to create CLIP embeddings for all LAION-400M images. We find that caption-image-concept similarity performs well for image classification accuracy. To support efficient multi-modal model training, we release the LAION-CLASS-Collected parquets for research use.

\subsection{Projector training details}
\label{sec:projector_training}
We use the standard CLIP loss with a learnable temperature parameter to train the projectors while keeping the vision and text encoders frozen. For our largest experiments on the 20M MIX-CLASS-Collected dataset, we use an effective batch size of 16k and train for 30 epochs. Training is done with a cosine learning rate scheduler, ramping up to 1e-3 in the first epoch. Additional hyperparameters are detailed in the table in the appendix. The training process takes 50 hours on a node with 8 A100 GPUs.

\subsection{0-shot Segmentation Evaluation}
\label{sec:seg-eval-details}
In DINOV2-ARL, we perform 0-shot segmentation by computing cosine similarities between each patch and all the ground truth classes and subsequently upscaling to the target size. Each patch is then classified into a corresponding class. Consistent with previous studies, the intersection over union (IoU) is computed solely for the foreground classes. In the zero-shot segmentation process of CLIP models, we employ a technique similar to \cite{zhou2022extract} to alleviate the opposite visualization problem in CLIP models \cite{li2023clip}. The patch embeddings from the penultimate layer are passed through the value layer and output MLP of the final self-attention block, followed by projection into the joint embedding space using the vision projector. Meanwhile, our DINOv2-ARL model considers patch embeddings projected into the joint embedding space by the patch projector and augments them with the projected CLS token in a residual manner.

\subsection{Multi-Lingual Full Results}
\label{sec:multi_full}
Another significant advantage of using only Projectors to align modalities is the ability to swap the text encoder with multi-lingual encoders trained on various languages, thus potentially extending a CLIP model to accommodate any language. This feature is particularly beneficial for low-resource languages. We demonstrate the feasibility of this approach by training projectors to align the DINOv2 visual encoder with the paraphrase-multilingual-v2 text encoder, using a dataset consisting solely of English image-caption pairs. We selected this specific text encoder as it showed the highest compatibility in terms of CKA with DINOv2. Subsequently, we evaluated the performance of our model on multi-lingual image retrieval using the XTD dataset \cite{aggarwal2020towards} and on multi-lingual image classification using the ImageNet dataset. For multi-lingual classification, we translate our VDT prompts \cite{Maniparambil_2023_ICCV} to the languages being considered using the nllb-700M model \cite{costa2022no} and then use the same prompts for all the models being considered including ours.

For both multi-lingual classification and retrieval tasks, our comparisons are structured into two categories as delineated in Table \ref{tab:multi_cls} and Table \ref{tab:multi_ret}. The lower sections of each of these tables list models trained exclusively with English captions, more specifically the CLIP-VIT-L models from OpenAI and LAION trained on 400 million image caption pairs of WIT dataset and LAION400M dataset respectively. The upper sections of these tables feature models trained with translated captions, including those employing contrastive training with multi-lingual image-caption pairs such as CLIP-models based on the LAION5B multi-lingual dataset, which contains image-caption pairs in over 100 languages. We also compare against, M-CLIP \cite{chen2023mclip} models that are trained using English and translated captions to align a multi-lingual text encoder with CLIP’s original text encoder through contrastive learning, thereby enhancing performance on multi-lingual tasks. Additionally we also compare against the NLLB-CLIP \cite{visheratin2023nllb} models developed through LiT \cite{zhai2022lit} techniques, coupling a frozen CLIP visual encoder with an unfrozen multi-lingual text encoder using translated captions from the smaller LAION-COCO dataset. We compare against only model sizes of up to ViT-Large for fair comparison.

\textbf{Retrieval results}: Our model DINOv2-MpNet trained only on English image,caption pairs outperforms all other CLIP models trained only on English image caption pairs, by a large margin of over 43 \(\%\) on average retrieval performance over 10 languages. We also outperform the next best performing English CLIP model trained on LAION400m English caption retrieval by over 6 percent. On Latin script languages the CLIP models have decent performance while it falls significantly for non Latin languages like JP, KO, PL, RU, TR, and ZH. This is mainly because these models were trained using an English only tokenizer which results in unknown token for most characters of these languages. However our DINOv2-MpNet projector model maintains competitive performance on all languages both Latin script and non Latin script even when compared against models specifically trained using multi-lingual data (Upper half of the table). Amongst the multi-lingual trained CLIP models we perform better than laion5b trained xlm-roberta-base-VitB32 by 4.5 percent. It is to be noted here that we only use 20 million Image caption pairs for alignment while LAION5B has over 5B image-caption pairs from over 100 languages and multi-lingual webli has over 30B image-caption pairs from over 100 languages. It is to be noted that our DINOv2-Mpnet is also competitive with M-CLIP model  XLM-Roberta-Large-Vit-B-16Plus(56.1 vs 57.7) which has been trained using translated English sentences of over 175 million data points to over 100 languages, and 3M translated image, caption pairs from CC3m.

\textbf{Classification results}: We see a similar trend when we compare our DINOv2-MpNet projector model against CLIP baselines(lower section), and multi-lingual baselines (upper section) on multi-lingual imagenet classification in Table. Our model showcases competitive performance to that of OpenAI-clip model while beating LAION400m trained ViT-Large on english Imagenet, while performing significantly better on all other languages considered (over 24 percent better on 8 language average). When compared with models trained with multi-lingual data, our model outperforms both nllb-clip models as well as M-CLIP models, beating the next best performing model M-CLIP/XLM-Roberta-Large-Vit-L-14 by over 3 percent despite not training using any multi-lingual text data. We believe that training using translated image-caption pairs of our dataset would further improve the performance of our method, and we leave this as a future work. The main advantage of training using our methods is that we can get highly porformant CLIP-like models using much lesser amount of image-caption pairs, (more than 20x lesser) resulting in quick adaptation to low resource languages given that a multi-lingual text encoder exists for that language.

\begin{table}
    \resizebox{\linewidth}{!}{%
    \begin{tabular}{lrrrrrrrrrrr|r}
    \toprule
    model & EN & DE & ES & FR & IT & JP & KO & PL & RU & TR & ZH & average \\
    \midrule
    nllb-clip-base@v1 & 47.2 & 43.3 & 44.1 & 45.0 & 44.7 & 37.9 & 39.4 & 45.5 & 40.6 & 41.2 & 41.1 & 42.3 \\
    M-CLIP/XLM-Roberta-Large-Vit-B-32 & 48.5 & 46.9 & 46.4 & 46.1 & 45.8 & 35.0 & 36.9 & 48.0 & 43.2 & 45.7 & 45.4 & 43.9 \\
    M-CLIP/XLM-Roberta-Large-Vit-L-14 & 56.3 & 52.2 & 52.7 & 51.8 & 53.6 & 41.5 & 42.5 & 54.1 & 48.4 & 52.7 & 53.5 & 50.3 \\
    xlm-roberta-base-ViT-B-32@laion5b & 63.2 & 54.5 & 54.6 & 55.7 & 55.7 & 47.1 & 43.8 & 55.5 & 50.3 & 48.2 & 50.8 & 51.6 \\
    nllb-clip-large@v1 & 59.9 & 56.5 & 56.7 & 56.0 & 55.5 & \textbf{49.3} & \textbf{51.7} & 57.4 & 50.4 & 56.0 & 52.3 & 54.2 \\
    M-CLIP/XLM-Roberta-Large-Vit-B-16Plus & 63.2 & \textbf{61.4} & \textbf{59.8} & 59.3 & \textbf{61.0} & 48.3 & 49.8 & 64.0 & 54.8 & \textbf{59.6} & \textbf{58.8} & \textbf{57.7} \\
    \midrule
    ViT-L-14@laion400m\_e31 & 64.5 & 26.7 & 31.4 & 38.3 & 26.6 & 1.4 & 0.4 & 4.8 & 1.7 & 4.1 & 1.0 & 13.6 \\
    openai/clip-vit-large-patch14 & 59.4 & 19.9 & 26.6 & 28.5 & 19.2 & 4.1 & 0.3 & 3.9 & 1.3 & 2.6 & 0.7 & 10.7 \\
    DINOv2-MpNet (Ours) & \textbf{70.7} & 60.6 & 59.0 & \textbf{60.6} & 60.7 & 45.6 & 49.8 & 58.3 & 52.7 & 55.8 & 57.9 & 56.1 \\
    \bottomrule
    \end{tabular}%
    
    }
    \setlength{\belowcaptionskip}{-10pt}
    \caption{\textbf{Multilingual image-caption retrieval} performance on XTD dataset. DINOv2-MpNet outperforms many baselines despite English-only training. Upper: multilingual-trained models; Lower: English-only trained models.}
    \label{tab:multi_ret}
\end{table}

\begin{table}
\resizebox{\linewidth}{!}{%
\begin{tabular}{lrrrrrrrr|r}
\toprule
model & EN & AR & ES & FR & DE & JP & ZH & RU & average \\
\midrule

nllb-clip-base@v1 & 25.4 & 20.4 & 23.9 & 23.9 & 23.3 & 21.7 & 20.3 & 23.0 & 22.4 \\
nllb-clip-large@v1 & 39.1 & 30.1 & 36.5 & 36.0 & 36.2 & 32.0 & 29.0 & 33.9 & 33.4 \\

M-CLIP/XLM-Roberta-Large-Vit-B-32 & 46.2 & 33.4 & 43.7 & 43.3 & 43.3 & 31.6 & 29.1 & 38.8 & 37.6 \\
M-CLIP/XLM-Roberta-Large-Vit-B-16Plus & 48.0 & 35.1 & 46.6 & 45.4 & 46.1 & 32.9 & 31.3 & 40.3 & 39.7 \\
xlm-roberta-base-ViT-B-32@laion5b & 63.0 & 29.0 & 53.4 & 53.8 & 55.8 & 37.3 & 26.8 & 40.3 & 42.3 \\
M-CLIP/XLM-Roberta-Large-Vit-L-14 & 54.7 & \textbf{40.0} & 51.9 & 51.6 & 51.9 & 37.2 & \textbf{35.2} & 47.4 & 45.0 \\
\midrule
ViT-L-14@laion400m\_e32 & 72.3 & 6.4 & 44.7 & 49.9 & 48.2 & 2.7 & 2.3 & 4.5 & 22.7 \\
openai/clip-vit-large-patch14 & \textbf{75.6} & 6.7 & 46.2 & 49.6 & 46.7 & 6.6 & 2.2 & 3.5 & 23.1 \\
DINOv2-MpNet (Ours) & 73.4 & 38.0 & 56.8 & 58.3 & 61.6 & \textbf{43.2} & 33.3 & 49.3 & 48.6 \\
\bottomrule
\end{tabular}
}
\caption{\textbf{Multi-lingual classification.} Classification performance comparison of DINOv2-MpNet and various CLIP models and multilingual baselines on multilingual ImageNet. Our DINOv2-MpNet model trained only on English data outperforms even models trained on multi-lingual data.
The upper half of the table lists models trained on multiple languages, while the lower half lists models trained only on English data. The models are evaluated on translations of the labels and the prompts made using nllb-200-distilled-600M translation model. \cite{costa2022no}}
\label{tab:multi_cls}
\end{table}

\subsection{Dataset scale}

\begin{figure}
    \centering
    \includegraphics[width=\linewidth]{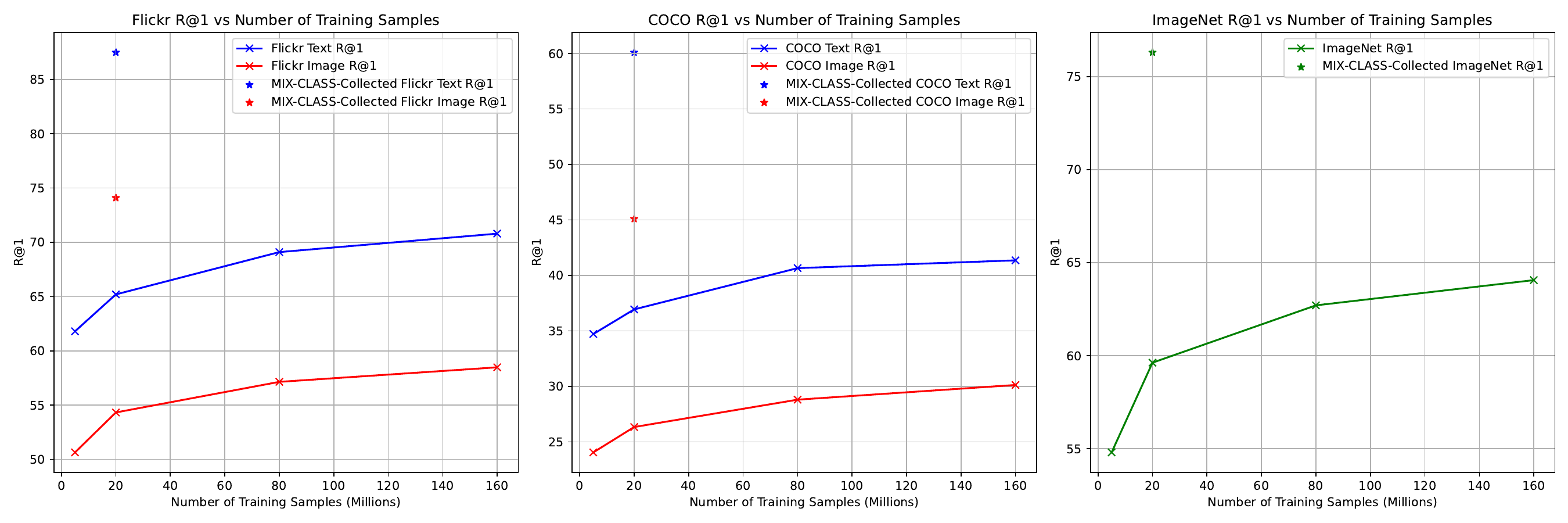}
    \caption{\textbf{Performance scales with higher amounts of randomly sampled LAION data} The performance scales with higher amounts of randomly sample data from LAION400M, but very slowly, highlighting the need for a densely covered and high quality dataset when training projectors only to align modalities.}
    \label{fig:dataset_scale}
\end{figure}

Figure \ref{fig:dataset_scale} illustrates that while performance scales with an increasing number of randomly sampled data points from the LAION400M dataset, the rate of improvement diminishes, highlighting the critical need for densely covered and high-quality datasets when training projectors to align modalities. Additionally, the comparative performance of MIX-CLASS-Collected data reveals that datasets curated with more focused criteria can lead to better performance gains than simply increasing the volume of data. This underscores the importance of prioritizing dataset quality over quantity, especially given the observed diminishing returns when using larger data sizes for projector-based alignment.

\subsection{sDCI benchmark results}
\label{sec:sdci}
We evaluate our method on the Densely Captioned Images (DCI) dataset \cite{urbanek2024picture}, which contains 7,805 images with mask-aligned descriptions averaging over 1,000 words each. To accommodate current models' token limits, the authors also provide sDCI, a summarized version with CLIP-compatible 77-token captions generated by LLMs.

sDCI introduces several benchmarks:

\begin{itemize}
    \item All SCM (Subcrop-Caption Matching): Matches captions to corresponding image subcrops.
    \item All Neg: Distinguishes between positive captions and LLM-generated negatives.
    \item All Pick5-SCM: Similar to All SCM, but uses multiple captions per subcrop.
    \item All Pick5-Neg: Distinguishes between multiple positive captions and a negative.
    \item Base Neg: Focuses on caption-negative distinction for full images only.
    \item All Hard-Negs: Uses the most challenging LLM-generated negatives.
\end{itemize}

We tested our DINOv2-ARL model on the sDCI dataset benchmarks. Table \ref{tab:sdci_results} presents our results alongside the CLip baseline. Our method demonstrates competitive performance compared to the CLIP baseline across several DCI benchmarks. 

\begin{table}

\resizebox{\columnwidth}{!}{%
\begin{tabular}{lcccccc}
\hline
Model & All SCM & All Neg & All Pick5-SCM & All Pick5-Neg & Base Neg & All Hard-Negs \\
\hline
CLIP Baseline & \textbf{40.06}\% & 60.79\% & \textbf{11.21}\% & \textbf{24.06}\% & 67.56\% & 41.34\% \\
DINOv2-ARL (Ours) & 29.33\% & \textbf{64.36}\% & 9.35\% & 21.39\% & \textbf{81.94}\% & \textbf{61.10}\% \\
\hline
\end{tabular}
}
\caption{Performance comparison on DCI dataset benchmarks}
\label{tab:sdci_results}
\end{table}

In the Subcrop-Caption Matching tasks (All SCM and All Pick5-SCM), our model performs slightly below the CLIP baseline. This suggests that there is room for improvement in our approach when it comes to distinguishing between the different parts that compose an image.

However, our model shows notable improvements in the negative detection tasks. We outperform CLIP on All Neg (64.36\% vs. 60.79\%), Base Neg (81.94\% vs. 67.56\%), and All Hard-Negs (61.10\% vs. 41.34\%). These results demonstrate the potential of our method in aligning vision and language models for a fine-grained understanding of image content, especially in scenarios requiring robust discrimination between relevant and irrelevant captions. Future work could focus on improving the model's performance on sub-crop caption matching tasks while maintaining its strong capabilities in negative detection.

\subsection{0-Shot Classification and Retrieval Evaluation Datasets}
\label{sec:eval_datasets}
To evaluate the performance of our DINOv2-ARL projector model and compare it with baseline CLIP models, we utilized a diverse set of datasets for zero-shot classification and retrieval tasks. These datasets span various domains and challenge the models' ability to generalize across different visual concepts.

For zero-shot classification, we employed the following datasets:

\begin{itemize}
    \item ImageNet \cite{imagenet}: A large-scale dataset with 1000 object categories, widely used as a benchmark for image classification tasks. It contains over 1.2 million training images and 50,000 validation images, with each image labeled with one of 1000 object classes.
    
    \item ImageNetV2 \cite{recht2019imagenet}: A newer version of ImageNet designed to test the robustness of models trained on the original ImageNet. It features 10,000 new test images collected using the same procedure as the original, but addressing certain biases in the original dataset.
    
    \item Caltech101 \cite{li_andreeto_ranzato_perona_2022}: A dataset containing pictures of objects belonging to 101 categories, plus a background category. It includes about 40 to 800 images per category, with most categories having about 50 images. The dataset is known for its high intra-class variability.
    
    \item Oxford-IIIT Pet \cite{parkhi2012cats}: A 37-category pet dataset with roughly 200 images for each class, featuring different breeds of cats and dogs. It includes pixel-level trimap segmentations and breed-level labels for each image.
    
    \item Stanford Cars \cite{krause20133d}: A dataset of 196 car classes, totaling 16,185 images. Classes are at the level of Make, Model, Year (e.g., 2012 Tesla Model S). It includes 8,144 training images and 8,041 testing images, with bounding box annotations.
    
    \item Oxford Flowers102 \cite{nilsback2008automated}: A 102 category dataset consisting of 102 flower categories common to the UK. It contains 40 to 258 images per class and provides segmentation data for each image. The dataset is particularly challenging due to the fine-grained nature of the categories.
    
    \item Food101 \cite{bossard2014food}: A large dataset of 101 food categories, with 101,000 images. It features 1000 images per food class, with 250 test images and 750 training images per class. The training images are not manually cleaned, adding a level of noise to the dataset.
    
    \item FGVC Aircraft \cite{maji2013fine}: A fine-grained visual classification dataset with 10,200 images of aircraft, spanning 100 aircraft models. Each model is associated with a specific variant, manufacturer, family, and collection. The dataset includes 6,667 training images and 3,333 test images.
    
    \item SUN397 \cite{rouach2020sun}: A scene recognition dataset with 397 categories and 108,754 images, covering a large variety of environmental scenes under various lighting conditions. It provides at least 100 images per class and has been used extensively for scene recognition tasks.
    
    \item Caltech-UCSD Birds-200-2011 (CUB) \cite{wah2011caltech}: A dataset for fine-grained image classification with 200 bird species, containing 11,788 images. Each image has detailed annotations including 15 part locations, 312 binary attributes, and 1 bounding box. It's widely used for fine-grained visual categorization research.
    
    \item UCF101 \cite{soomro2012ucf101}: An action recognition dataset with 101 action categories, consisting of realistic action videos collected from YouTube. It contains 13,320 videos from 101 action categories, with videos exhibiting large variations in camera motion, object appearance and pose, illumination conditions, and more.
\end{itemize}

For zero-shot image-text retrieval, we used:

\begin{itemize}
    \item Flickr30k \cite{plummer2015flickr30k}: A dataset containing 31,783 images collected from Flickr, each paired with 5 crowd-sourced captions. It focuses on describing the objects and actions in everyday scenes. The dataset is split into 29,783 training images, 1000 validation images, and 1000 test images.
    
    \item COCO \cite{lin2014microsoft}: A large-scale dataset for object detection, segmentation, and captioning, which we use for its image-caption pairs in the retrieval task. It features over 330,000 images, each with 5 captions. The dataset includes 80 object categories and instance segmentation masks, making it versatile for various computer vision tasks.
\end{itemize}

These datasets comprehensively evaluate a model's ability to perform zero-shot classification across various domains and its capacity for cross-modal retrieval. By using this diverse set of benchmarks, we can assess the generalization capabilities of our approach compared to existing CLIP models. We use Visually Descriptive Class-Wise prompts from \cite{Maniparambil_2023_ICCV} to enable the unimodal-text encoder in our DINOv2-ARL projector model to better identify the zero-shot classes of the downstream datasets.

\subsubsection{Concept Coverage Collection datasets}
\label{subsubsec:concept_datasets}
We use a few shot examples from  14 curated computer vision datasets to construct our Concept Image prototypes to curate the images from our uncurated data pool. The 14 curated datasets are described as follows.

\begin{itemize}
    
\item BirdSnap \cite{berg2014birdsnap}: A fine-grained dataset consisting of 49,829 images of 500 North American bird species. The images are annotated with species labels, and the dataset is primarily used for species classification and fine-grained recognition tasks.

\item Caltech101 \cite{li_andreeto_ranzato_perona_2022}: A dataset containing pictures of objects belonging to 101 categories, plus a background category. It includes about 40 to 800 images per category, with most categories having about 50 images. The dataset is known for its high intra-class variability.

\item EuroSAT \cite{helber2019eurosat}: A satellite image dataset with 10 categories related to land use classification (e.g., forests, rivers, residential areas). It contains 27,000 labeled images, with 2700 images per class, widely used in remote sensing and geospatial tasks.

\item FGVC Aircraft \cite{maji2013fine}: A fine-grained classification dataset with 10,000 images of 100 aircraft model variants from 70 manufacturers. It is used for distinguishing between visually similar objects in fine-grained recognition tasks.

\item Flowers102 \cite{nilsback2008automated}: A dataset containing 102 flower categories, commonly used for fine-grained classification tasks. It has a total of 8,189 images, with 40 to 258 images per category, and is organized into a training, validation, and test set.

\item Food101 \cite{bossard2014food}: A dataset containing 101,000 images of 101 food categories. Each category has 750 training images and 250 test images, commonly used for food classification and recognition tasks.

\item GTSRB \cite{stallkamp2012man}: The German Traffic Sign Recognition Benchmark dataset, containing over 50,000 images of 43 different traffic sign classes. It is designed for multi-class classification tasks in the context of traffic sign recognition.

\item ImageNet \cite{imagenet}: A large-scale dataset with 1,000 object categories, widely used as a benchmark for image classification tasks. It contains over 1.2 million training images and 50,000 validation images, with each image labeled with one of 1,000 object classes.

\item Oxford Pets \cite{parkhi2012cats}: A dataset of 7,349 images, containing 37 categories of pets (both cats and dogs). Each image is annotated with species and breed information, commonly used for image classification and segmentation tasks.

\item RESISC45 \cite{cheng2017remote}: A dataset of remote sensing images used for scene classification, containing 31,500 images across 45 scene classes. Each class has 700 images with variations in resolution, scale, and orientation.

\item Stanford Cars \cite{krause20133d}: A dataset with 16,185 images of 196 car models, annotated by make, model, and year. The dataset is designed for fine-grained classification and recognition tasks of vehicles.

\item Pascal VOC 2007 \cite{everingham2015pascal}: A dataset for object detection, segmentation, and classification, containing 9,963 images of 20 object categories. It is widely used for benchmarking models in computer vision tasks.

\item SUN397 \cite{rouach2020sun}: A large-scale scene understanding dataset with 397 categories and 108,754 images. It covers a wide range of environments, from natural to man-made scenes, commonly used for scene classification tasks.

\item UCF101 \cite{soomro2012ucf101}: A video dataset consisting of 13,320 videos across 101 human action categories. It is widely used for action recognition tasks in video analysis and computer vision research.
\end{itemize}

\end{document}